\documentclass[journal,transmag]{IEEEtran}

\usepackage [table]{xcolor}
\usepackage[cmex10]{amsmath}

\usepackage[nodisplayskipstretch]{setspace}

\usepackage{graphics} 
\usepackage{epsfig} 
\usepackage{epstopdf}
\usepackage{subfig}
\usepackage{mathptmx} 
\usepackage{times} 
\usepackage{amsmath} 
\usepackage{amssymb}  
\usepackage{algorithm}
\usepackage{algpseudocode}
\usepackage{array}
\usepackage{bm}
\usepackage{xcolor}
\usepackage{float}
\usepackage{multirow}
\usepackage[font=small]{caption}
\usepackage{siunitx}
\usepackage{caption}
\usepackage{ctable}
\usepackage[colorlinks]{hyperref}
\usepackage{array}
\usepackage{tensor}
\usepackage{color, soul}
\renewcommand\abstractname{Abstract}
\usepackage{amssymb}

\ifCLASSOPTIONcompsoc
  \usepackage[nocompress]{cite}
\else
  \usepackage{cite}
\fi

\newcolumntype{P}[1]{>{\centering\arraybackslash}p{#1}}

\begin{document}
\title{A Multi-Robot Cooperation Framework for Sewing Personalized Stent Grafts}

\author{Bidan Huang,
        Menglong Ye,
        Yang Hu,
        Alessandro Vandini,
        Su-Lin Lee,
        Guang-Zhong~Yang,~\IEEEmembership{Fellow,~IEEE}
\thanks{B. Huang,  Y. Hu, S.-L. Lee and G.-Z. Yang are with the Hamlyn Centre for Robotic Surgery, Imperial College London, SW7 2AZ, London, UK (e-mail: b.huang@imperial.ac.uk; y.hu12@imperial.ac.uk; su-lin.lee@imperial.ac.uk; g.z.yang@imperial.ac.uk). \newline   
\indent M. Ye was with the Hamlyn Centre when he contributed to this paper and is now with the Auris Surgical Robotics Inc., San Carlos, USA  (e-mail: menglong.ye11@imperial.ac.uk). \newline 
\indent A. Vandini was with the Hamlyn Centre when he contributed to this paper and is now with the Samsung R\&D Institute, London, UK  (e-mail: a.vandini12@imperial.ac.uk).\newline
\indent The project is supported by the EPSRC (EP/L020688/1)
}
}

\maketitle

{%
\begin{abstract}

This paper presents a multi-robot system for manufacturing personalized medical stent grafts. The proposed system adopts a modular design, which includes: a (personalized) mandrel module, a bimanual sewing module, and a vision module. The mandrel module incorporates the personalized geometry of patients, while the bimanual sewing module adopts a learning-by-demonstration approach to transfer human hand-sewing skills to the robots. The human demonstrations were firstly observed by the vision module and then encoded using a statistical model to generate the reference motion trajectories. During autonomous robot sewing, the vision module plays the role of coordinating multi-robot collaboration. Experiment results show that the robots can adapt to generalized stent designs. The proposed system can also be used for other manipulation tasks, especially for flexible production of customized products and where bimanual or multi-robot cooperation is required.
\end{abstract}



}

{\section{Introduction}\label{sec:introduction}}

The latest development of robotics, sensing and information technology is driving the future of Industry 4.0. One of the key concepts is ``smart factories'': factories with flexible production lines to produce quality products tailored for customer requirements~\cite{russmann2015industry}. Different from the current practice, a smart factory requires a single production process to finish multiple products with different designs. Such a scheme needs to be equipped with: 1) smart machines or robots that can program themselves automatically according to customized designs; 2) self-optimizing mechanisms that can react to changing conditions during the production process; 3) real time sensing ability to monitor and guide the process to ensure accuracy.

In this paper, we propose a multi-robot manufacturing scheme for flexible production of personalized medical stent grafts. A stent graft, as shown in Fig.~\ref{fig:sewinglady}, is a tubular structure composed of a fabric tube, the graft, supported by multiple metal rings called stents. It is a medical device commonly used during endovascular surgery for treating vascular diseases such as Abdominal Aortic Aneurysms (AAA), a major contributor to cardiovascular related deaths in the Western world~\cite{cdc2013}.

While endovascular repair has been increasingly utilized, there are very few providers of personalized stent grafts~\cite{custom2016}. Each personalized stent graft is designed to fit a patient's specific anatomical structures, e.g. the diameter and length of the aneurysm, obtained from their computed tomography (CT) or magnetic resonance (MR) scans. Similar to tailored garments, most of these personalized devices are handmade, requiring an extensive period of human crafting. The current process can take weeks or even months, subjecting patients to significant risks of deadly aneurysm rupture. Autonomous manufacturing for custom-made stent grafts therefore provides tremendous potential and is an unmet clinical demand.

The production of personalized stent grafts is service-oriented: custom-designed products are based on customer demands. To this end,
the proposed robotic system incorporates personalization, learning, and adaptation combined with real-time vision. It leverages the latest additive manufacturing processes to make patient-specific models. To reduce the cost associated with manufacturing customized products, we adopted a modular design to separate the repetitive tasks from the personalized tasks. The entire process is monitored under a real time vision system for multi-robot coordination.

\vspace{1cm}
\section{Related Work}\label{sec:relatedwork}

\begin{figure}
\centering

\subfloat[\scriptsize{}]{\includegraphics[height=3.9cm]{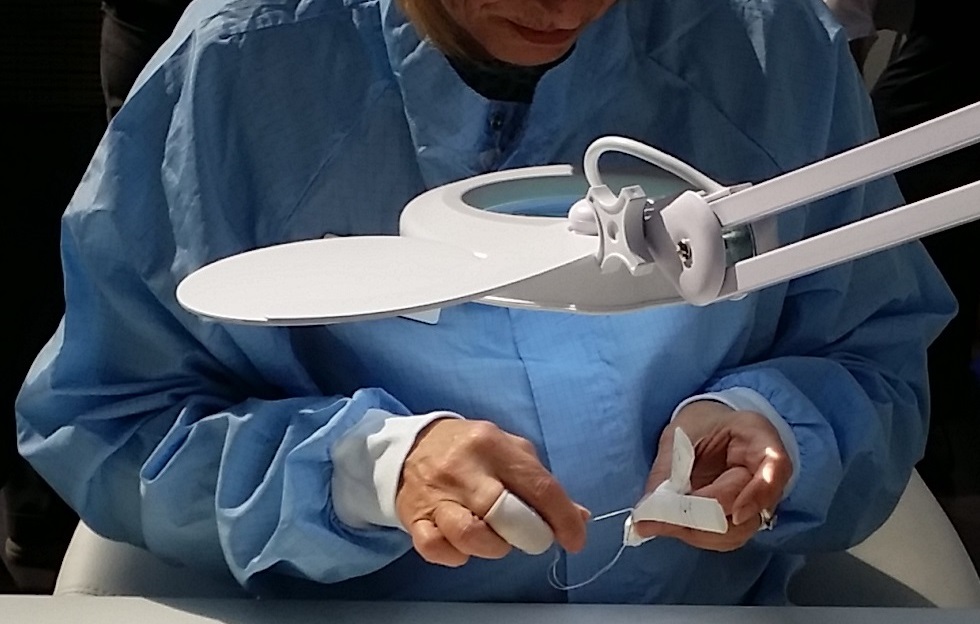}}
\subfloat[\scriptsize{}] {\includegraphics[height=3.9cm]{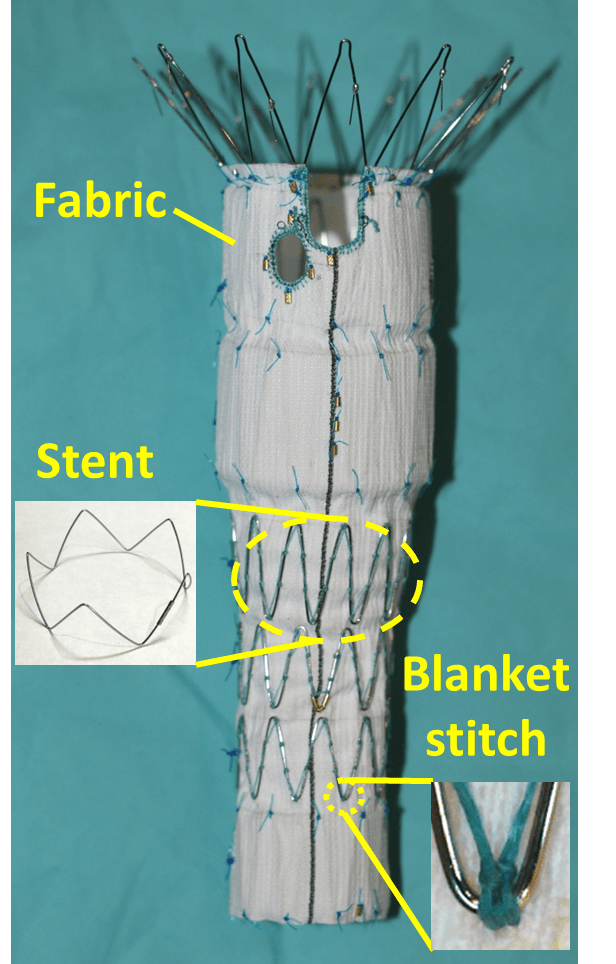}}
\caption{(a) Hand sewing of a personalized stent graft. (b) A personalized stent graft. The maximum diameter is 34 $mm$.}
\label{fig:sewinglady}
\end{figure}

Research into automated personalized manufacturing has received extensive interest in recent years. Most studies focus on the high level design of the system, such as incorporating wireless networks and cloud computing in factories~\cite{wang2016implementing}, using big data for product modelling and customization~\cite{wang2016towards}, and automating the supply network~\cite{montreuil2005demand}. Less studied is how to automate a production cell/line to produce customized products ~\cite{liao2017past}.

Extensive research on automated sewing has also been performed in the textile industry. Most of the existing research has been focused on augmenting the capability and intelligence of conventional sewing machines. Relevant topics in this field include multi-arm robotic sewing~\cite{kudo2000multi}, tension control for fabric, and edge tracking~\cite{schrimpf2012experiments}. To cope with environmental changes during sewing, various control strategies have been implemented, such as fuzzy logic controllers~\cite{koustoumpardis2006intelligent}, hybrid position/force control~\cite{kudo2000multi}, and leader/follower control strategies~\cite{schrimpf2014velocity}. Research has also been carried out on the redesign of sewing machines that are capable of sewing from a single side of the fabric and therefore facilitate sewing on complex 3D surfaces.
For example, KSL Keilmann (Lorsch, Germany) has developed different single sided sewing heads for 3D sewing fabric-reinforced structures for aircraft parts. These sewing systems, however, are designed to sew large and heavy objects. Single sided sewing of delicate objects remains a challenging problem.

\begin{figure}
\centering
\includegraphics[width=9cm]{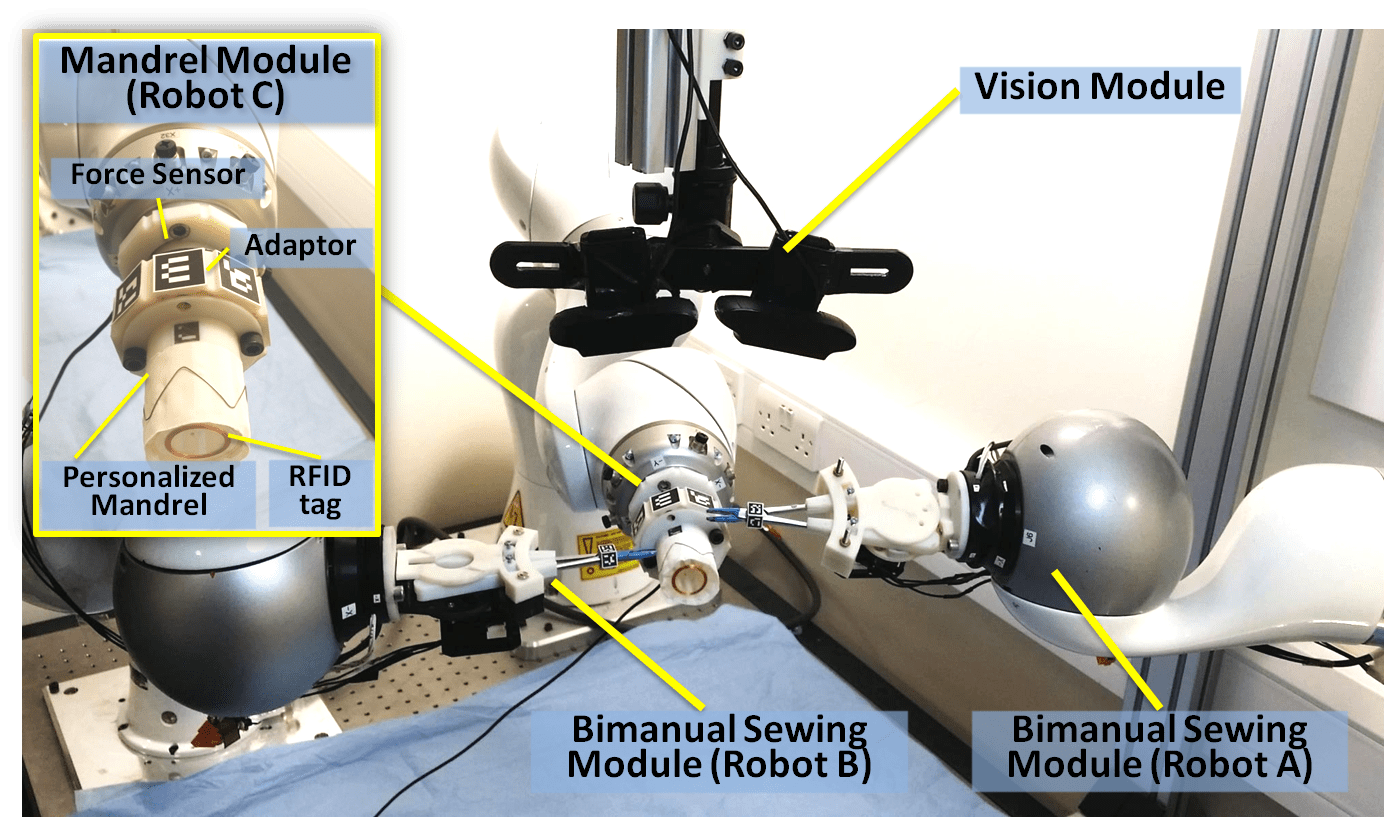}
\caption{The modularized multi-robot system for stent grafts sewing. }
\label{fig:setup}
\end{figure}

Medical devices such as stent grafts demand personalization and creating products based on the customers' specifications is the main motivation of Industry 4.0.
In this paper, we propose a solution for a high-value medical device manufacturing challenge: how to enable robots to produce a variety of customized products with low costs.
The proposed system uses personalized stent graft sewing as the exemplar to demonstrate a solution. This work focuses on the task that is most challenging to automate: sewing the stents on the fabric. This task involves dexterous manipulation of the needle, fabric, and thread. The personalized stent graft is currently hand-crafted with manually sewn stents (Fig.~\ref{fig:sewinglady}). Despite the speed of conventional sewing machines, they lack the ability to adapt to different free-form 3D geometries.


As bimanual sewing involves intricate coordinated motions, we have adopted a learning by demonstration approach.
Existing learning by human demonstration methods range from the simple “record-and-replay” method to more sophisticated approaches of incorporating visual servoing to cater for positional variation and different poses of grasping~\cite{pan2012recent}.
However, these approaches are mainly used to demonstrate tasks for a single robot arm. Demonstration of a bimanual task is difficult to achieve by a single user. Delicate bimanual tasks such as sewing and surgical suturing are largely demonstrated via tele-operation~\cite{van2010superhuman}, which requires an extra master-slave system. We propose herewith an efficient demonstration method to program the sewing motion via a vision system. The merits of this method are twofold: 1) users can demonstrate the task accurately and intuitively with their own hands rather than via the control panel or via kinaesthetic teaching and 2) the demonstration does not require the robots and hence can be done without interrupting the current robotic production process (detailed in Section~\ref{sec:overview:bimanual}).

However, the effectiveness of vision-based manipulation also relies on the accuracy of tool tracking and detection~\cite{Ye2016}. In medical suturing, small objects such as needles are difficult to track by camera. To this end, Iyer et al.~\cite{iyer2013single} proposed a single-camera system for auto-suturing with a monocular pose measurement algorithm~\cite{lo2002trip}. A 3D stereo system was proposed~\cite{staub2010automation} to improve the accuracy of aligning the needle with the target stitching point. The ``look-and-move'' visual servoing method~\cite{hutchinson1996tutorial} can be used to increase task accuracy and compensates the kinematic errors of the robots. In this paper, we present a robust needle detection algorithm. Fig.~\ref{fig:setup} illustrates the major components of the proposed manufacturing system. The main contributions of this work include:

\begin{enumerate}
\item {A modular multi-robot system that enables flexible production of customized medical devices;}
\item {A novel hardware design (mandrel) to cater for personalized product design;}
\item {An easy-to-use method for users to demonstrate intricate bimanual tasks;}
\item {A vision-based system for communication between multiple robots and visual servoing.}
\end{enumerate}

\section{Overview}
\label{sec:overview}

\begin{figure}
\subfloat[\scriptsize{}]{\includegraphics[height=3.9cm]{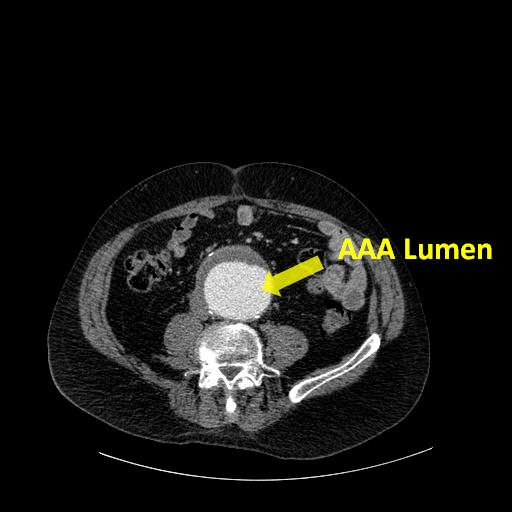}}
\subfloat[\scriptsize{}]{\includegraphics[height=3.9cm]{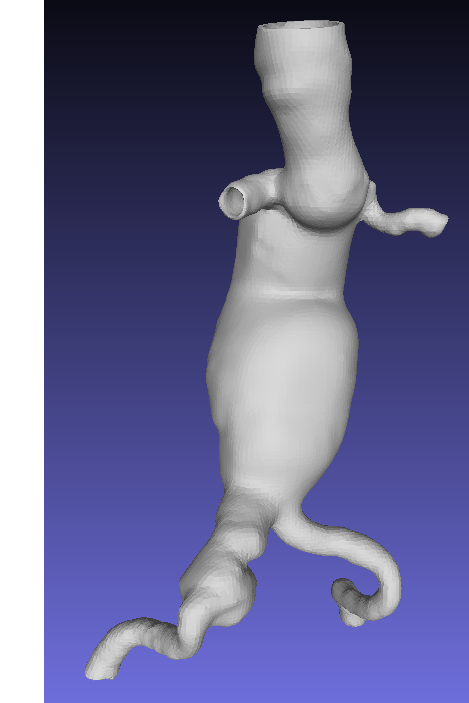}}
\subfloat[\scriptsize{}]{\includegraphics[height=3.9cm]{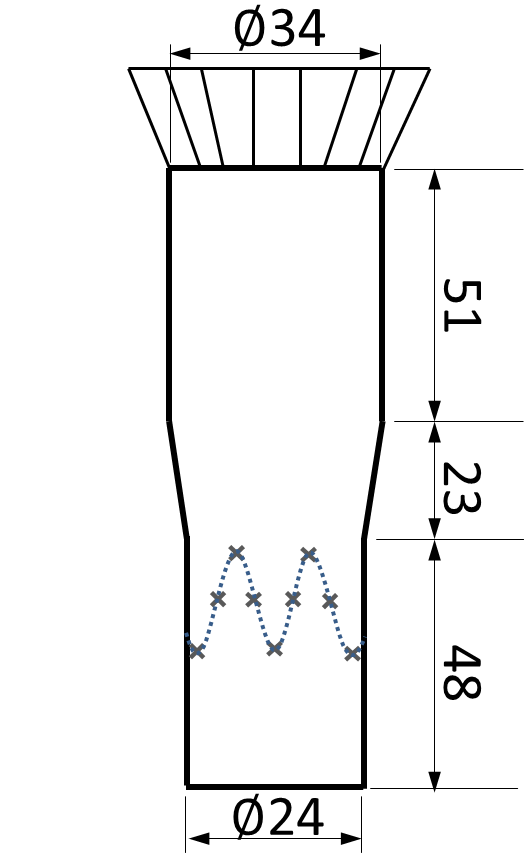}}
\caption{Three stages of designing a personalized stent graft. (a) The patient's CT scan. (b) A 3D reconstruction of the patient's AAA. (c) A sketch of the patient specific designed stent graft. The numbers are in $mm$. The dotted curve represents the location of one of the stent rings and the crosses represent the locations of stitches for the stent.}
\label{fig:designstent}
\end{figure}

The multi-robot sewing system is designed with a modular scheme. As shown in Fig.~\ref{fig:setup}, our proposed system is composed of three modules: the bimanual sewing module (Kuka Robots A, B, Needle Drivers A, B), the personalized mandrel module (Robot C, Force sensor\footnote{Optoforce OMD-20-FE-200N}, Mandrel, Fabric), and the vision module (stereo camera). The sewing task is hence separated into two parts: bimanual sewing and handling the personalized mandrel. The bimanual sewing motion is learned by observing human demonstrations (Section~\ref{sec:overview:bimanual}) while the mandrel motion is computed according to the patient-specific design (Section~\ref{sec:overview:mandrel}). The three robots are coordinated via the vision module (Section~\ref{sec:overview:vision}).

With this modular design, we encapsulate the customized components of the system into a single module, i.e. the mandrel module. The mandrel determines not only the patient specific geometry but also the exact position of each stitch. When sewing a new stent graft design, only the mandrel module needs to be reconfigured whilst all other modules remain the same. Hence, the cost of reconfiguring the system for customization is minimized.

Specifically, the role of each module is listed as below:

\begin{enumerate}
\item Mandrel module
\begin{enumerate}
\item Personalized trajectory planning
\item Handling the stents and fabric
\item Monitoring thread tension
\end{enumerate}
\item Bimanual sewing module
\begin{enumerate}
\item Learning and reproducing human hand sewing
\item Real time adaptation
\end{enumerate}
\item Vision module
\begin{enumerate}
\item Watching and recording user demonstrations of bimanual sewing
\item Coordinating robots, tracking, and visual servoing
\item Detecting the needle pose
\end{enumerate}
\end{enumerate}

\subsection{Mandrel Module}
\label{sec:overview:mandrel}
\begin{figure}
\centering
{
\subfloat[\scriptsize{Design of a mandrel}] {\includegraphics[height=3.3cm]{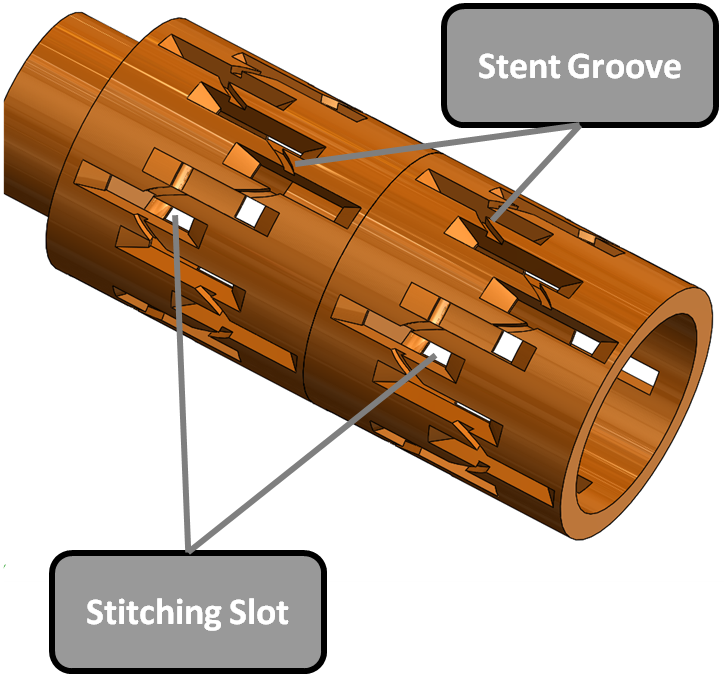}}
\hspace{0.5cm}
\subfloat[\scriptsize{Design of a needle driver}] {\includegraphics[height=3.3cm]{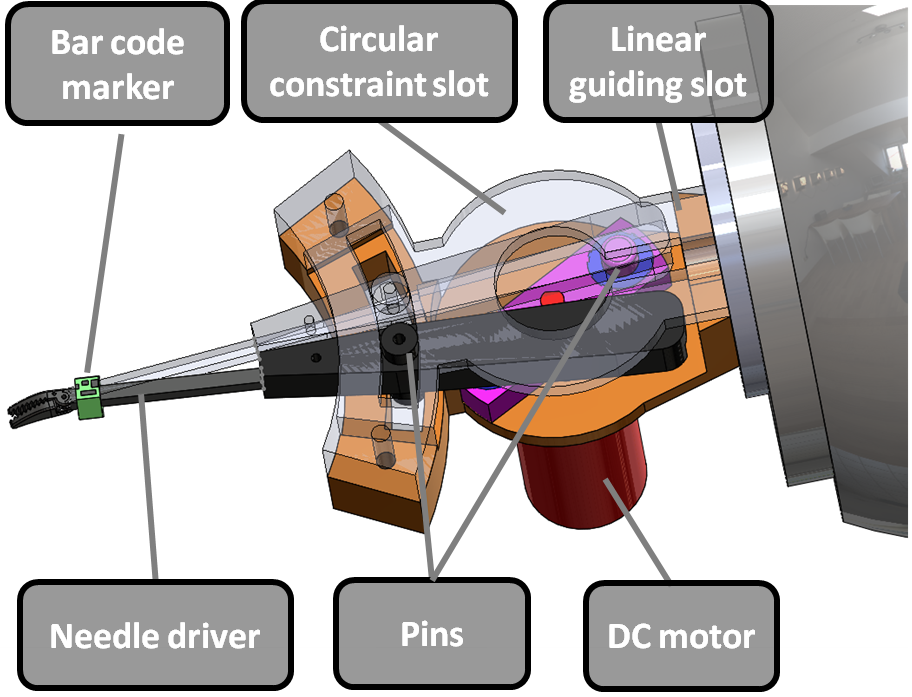}}
\caption{(a) The mandrel. The stitching slots allow the needle to sew and the grooves are for fixing the stents. (b) The motorised needle driver. It is designed to be attached to the Kuka robot and has a DC motor that opens and closes it. }
\label{fig:hardware}
}
\end{figure}

The key to the personalization of a stent graft is the 3D printed patient specific mandrel. A mandrel is a hollow cylinder (or cylinder-like shape) to support the fabric and the stents. Fig.~\ref{fig:hardware}a shows a basic mandrel. This mandrel serves two important roles during the manufacturing process: 1) tightly binding the stents and the fabric together and 2) enabling the robots to sew in the correct locations.

The shape of the mandrel is customized together with the stent graft to fit to each patient's anatomy. Fig.~\ref{fig:designstent} illustrates the three stages of designing a stent graft and the corresponding mandrel. Starting from the patient's CT/MR scan images (Fig.~\ref{fig:designstent}a), the 3D geometry of the aorta and the aneurysm is reconstructed (Fig.~\ref{fig:designstent}b). A stent graft is hence designed based on this 3D model and the mandrel is designed to be in the same shape, with grooves arranged for positioning the stents and sewing slots for needle piercing. The design specification and the CAD model are stored in a shared repository.
This information can be retrieved, for example, via a radio-frequency identification (RFID) tag attached to the 3D printed mandrel (Fig.~\ref{fig:setup}).

Prior to sewing, the mandrel, which is wrapped with the graft fabric and stents, is affixed to a 3D printed adaptor (Fig.~\ref{fig:setup}). This adaptor is an octagonal prism with a vision-based marker attached to each face. The pose of the mandrel is computed by detecting the pose of these markers (Section~\ref{sec:overview:vision}). A force sensor for monitoring the thread tension (Section~\ref{sec:experiment}) is affixed to the other side of the adaptor.
This mandrel-adaptor-force sensor setup is then mounted to the robot end-effector. Installing a new mandrel is relatively simple and takes two minutes on average.

The mandrel movement controlled by Robot C is computed according to the mandrel's design, i.e., the location of each stitching slot. After the mandrel design is retrieved,  Robot C plans its motion trajectory automatically and delivers the first stitching slot to the initial location to start sewing. Upon completion of each stitch, Robot C moves the mandrel to allow the bimanual sewing module to access the next stitching slot easily. In this way, the system is adaptable to different personalized stent graft designs and the customization is achieved by simply changing the mandrel.


\subsection{Bimanual Sewing Module}
\label{sec:overview:bimanual}

\begin{figure*}
\centering
{
\subfloat[\scriptsize{}] {\includegraphics[height=2.2cm]{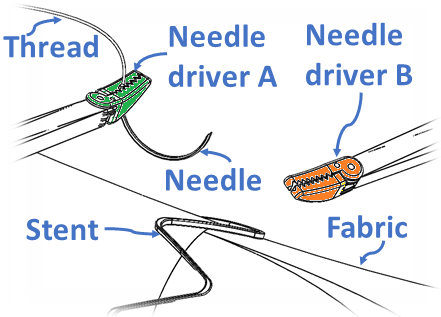}}
\subfloat[\scriptsize{}] {\includegraphics[height=2.2cm]{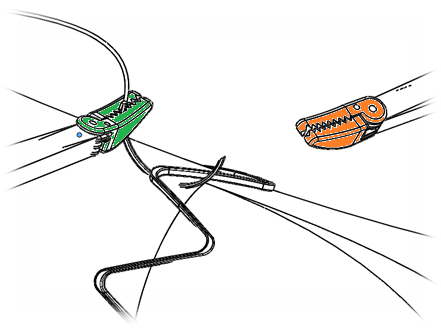}}
\subfloat[\scriptsize{}] {\includegraphics[height=2.2cm]{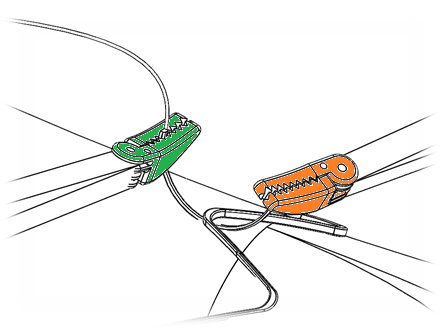}}
\subfloat[\scriptsize{}] {\includegraphics[height=2.2cm]{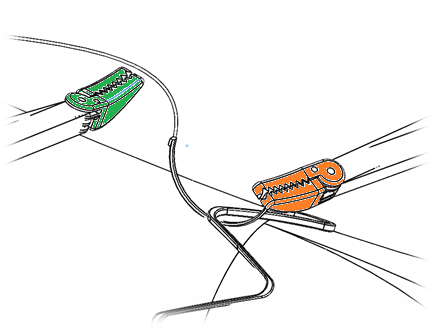}}
\subfloat[\scriptsize{}] {\includegraphics[height=2.2cm]{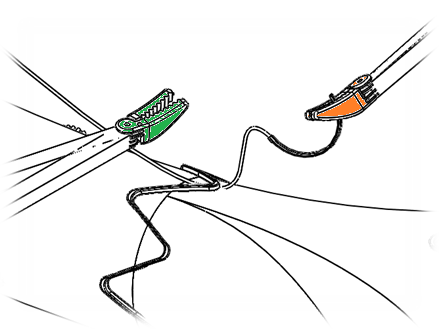}}
\subfloat[\scriptsize{}] {\includegraphics[height=2.2cm]{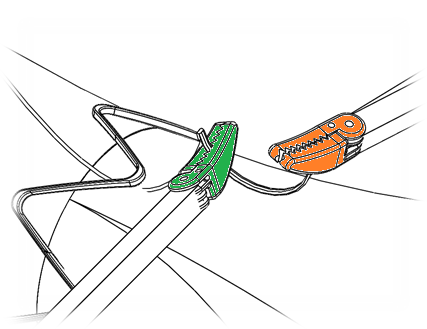}}
\caption{The key steps involved in one stitch cycle (a-f), of which at the end the needle is passed back to Needle Driver A for the next cycle.}
\label{fig:stitchcycle}}
\end{figure*}

The bimanual sewing module manages the key motion element: sewing. Hand sewing is a laborious job requiring fine manipulation skills. Although there exists a large variety of specialized sewing machines, many hand-sewing tasks are difficult to automate. To this end, we have adopted a learning by human demonstration approach to simplify the task.

Intricate hand motion was first demonstrated by a user and then segmented into multiple motion primitives (Section~\ref{sec:overview:data}), each encoded with a statistical model (Section~\ref{sec:overview:learning}). These models were then implemented to allow the robot to reproduce the same stitches (Section~\ref{sec:overview:context}). The demonstrations were observed by the vision system and the learned reference trajectory was later reproduced under vision guidance and servoing. This approach is generic to position-based bimanual tasks and hence is applicable to similar flexible production lines/cells.

Two robot arms (Kuka Robots A and B) mounted with two end effectors (Needle Driver A and B, as shown in Fig.~\ref{fig:hardware}b)\footnote{These are surgical needle drivers that are specially designed to grip the needle firmly. They are motorized to allow the robots to open and close them.} and a curved needle were used for bimanual sewing. This is a typical bimanual manipulation system and the target object is the needle. In subsequent sections, we refer to the needle as the \textit{object}, and the needle drivers as the \textit{tool}. Here we use an “object centric” approach: the human skill is represented by the motion of the tools and the objects. During demonstration, the user controls the tools to manipulate the object. The motion of the tools and object, rather than the human hands, is recorded and learned. In task reproduction, the robots use the same tools to manipulate the same object. In this way, the human manipulation skills can be easily transferred to robots without the need of mapping the human motion to the robots.

\subsubsection{Data Acquisition}
\label{sec:overview:data}
The user demonstrates the task by using two needle drivers, with a bar-code marker mounted  on each for visual tracking. The vision module was mounted on top of the workspace to record the 6 d.o.f poses of the needle drivers. The sewing was performed on a pre-installed mandrel.

Sewing is a repetitive task and Fig.~\ref{fig:stitchcycle} shows the main steps of a single stitch cycle. A surgical 1/2 circular needle was used, of which the trajectory was computed according to the pose of the needle drivers (Section~\ref{sec:overview:needle}). When a needle driver was holding the needle, the trajectories of both the needle and the needle driver were recorded in the reference frame of the mandrel; when the needle driver was not holding the needle, only its own trajectory was recorded in the reference frame of the needle. Needle pose estimation was performed at the beginning of each stitching cycle. The user demonstrated the stitching process to the system multiple times to generate the training data.

\subsubsection{Task Learning}
\label{sec:overview:learning}
After low-pass filtering of the raw data, each demonstration was segmented to a series of motion primitives, according to the needle drivers' open and closed status and their attachment to the needle. These motion primitives are listed in Table~\ref{tab:segments}. Dynamic Time Warping was then applied~\cite{berndt1994using} to each primitive to temporally align all the trials.

\begin{table}
\fontsize{8}{12}\selectfont
\centering
\caption{Motion primitives of stitching}
\vspace{5pt}
\label{tab:segments}
\hspace{-0.5cm}
\begin{tabular}{P{1cm}P{0.8cm}P{1.5cm}P{1.5cm}P{1cm}}
\hline
\rowcolor[HTML]{C0C0C0}
\begin{tabular}[c]{@{}c@{}}Motion\\Primitives\end{tabular} & \begin{tabular}[c]{@{}c@{}}Steps in \\ Fig.~\ref{fig:stitchcycle}\end{tabular} & \begin{tabular}[c]{@{}c@{}}Needle Driver\\A status\end{tabular} & \begin{tabular}[c]{@{}c@{}}Needle Driver\\B status\end{tabular} & \begin{tabular}[c]{@{}c@{}}Needle\\status\end{tabular} \\ \hline \hline
1. & a, b & Closed & Open & With A \\ \hline
2. & c & Closed & Closed & With A \\ \hline
3. & d,e & Open & Closed & With B \\ \hline
4. & f & Closed & Closed & With B \\ \hline
5. & a & Closed & Open & With A \\ \hline
\end{tabular}
\end{table}

Each motion primitive was encoded by a 7D Gaussian Mixture Model (GMM) $\Omega$ and formed a motion primitive~\cite{calinon2007learning,Huang2013}. The time stamp $t$ and the 6 d.o.f pose $h = \{x,y,z,\alpha,\beta,\theta\}$ were encoded. The probability that a given point ${t,h}$ belongs to $\Omega$ is computed as the sum of the weighted probability of the point belonging to each Gaussian component ${\Omega}_k$:

\begin{equation}
\setstretch{1.5}
p\left(t,h\mid\Omega\right) = \sum_{k=1}^K \pi_k p_k\left(t,h\mid\mu_k,\Sigma_k\right)
\end{equation}
where $\pi_k$, $p_k$ are the prior and corresponding conditional probability density of the $k$-th Gaussian component (${\Omega}_k$), with ${\mu}_k$ and ${\Sigma}_k$ as mean and covariance. More specifically, the mean $\mu_k$ and the covariance $\Sigma_k$ are defined as:

\begin{equation}
\setstretch{1.5}
{
{\mu}_k = \begin{pmatrix} {\mu}_{t,k} \\
{\mu}_{h,k}
\end{pmatrix}
\hspace{0.2in}
{\Sigma}_k = \begin{pmatrix} {\Sigma}_{tt,k} & {\Sigma}_{th,k} \\
{\Sigma}_{ht,k} & {\Sigma}_{hh,k}
\end{pmatrix}
}
\end{equation}

To determine the number of Gaussian components $K$, a five-fold cross validation was used.

The reference trajectory of each motion primitive was retrieved via GMR by querying the mean $\hat{\mu}_h$ and the covariance $\hat{\Sigma}_{hh}$ of the pose at each time step $\hat{t}$:

\begin{equation}
\setstretch{1.5}
{
\hat{\mu}_{h} = \sum_{k=1}^K{\beta_k}\left(\hat{t}\right)\hat{\mu}_{h,k}
}
\hspace{1cm}
{
\hat{\Sigma}_{hh} = \sum_{k=1}^K{\beta_k}\left(\hat{t}\right)^2\hat{\Sigma}_{hh,k}
}
\end{equation}

where
\begin{equation}
\setstretch{1.5}
{
\hat{\mu}_{h,k} = {\mu}_{h,k} + \Sigma_{ht,k}({\Sigma}_{tt,k})^{-1}(\hat{t}-{\mu}_{t,k})
}
\vspace{-2mm}
\end{equation}

\begin{equation}
\setstretch{1.5}
{
\hat{\Sigma}_{hh,k} = {\Sigma}_{hh,k} - {\Sigma}_{ht,k}({\Sigma}_{tt,k})^{-1}{\Sigma}_{th,k}
}
\end{equation}
and
\begin{equation}
\setstretch{1.5}
{
\beta_k\left(\hat{t}\right) = \frac{\pi_{k}p(\hat{t}|{\mu}_{t,k},{\Sigma}_{tt,k})}
{\sum_{k=1}^K{\pi_k}p(\hat{t}|{\mu}_{t,k},{\Sigma}_{tt,k})}
}
\end{equation}

\subsubsection{Trajectory Optimisation for Task Contexts}
\label{sec:overview:context}
The learned reference trajectory needs to be further optimized to maximize the task performance in terms of both accuracy and speed. We achieved this by varying the speed of the task reproduction in different task contexts. Generally, a manipulation task has two task contexts: \textit{end point driven} and \textit{contact driven}.


In the bimanual sewing task, the approaching and exiting motion of the needle to the fabric is \textit{end-point driven}, while the piercing in and out motion belongs to the \textit{contact driven} context. To ensure stitch quality, the needle piercing in and out motion must follow the reference trajectory accurately. Hence in this context, the robot is slowed down to allow it to follow the reference trajectory carefully. For \textit{end-point driven} motion, however, the robot does not need to follow the exact trajectory, as long as it reaches the final destination.

To identify the task context, the variance between different demonstrations in each motion primitive was analyzed. As shown in Fig.~\ref{fig:demo}, the variance of the demonstrations varies across the task. Those parts with large variance were identified as end point driven, while those with small variance were identified as contact driven.

According to the bimanual sewing task requirement, we chose the correlation of the variance and the ratio to the demonstration speed $R$ as:
\[
R =
\begin{cases}
0.5,& var_t > 0.01 \text{ $or$ } var_r > 15\\
1.5 & var_t \in [0.005, 0.01] \text{ $or$ } \\
& var_r \in [5, 15]\\
2 & var_t<0.005 \text{ $or$ } var_r<5\\
\end{cases}
\]
where $var_t$ and $var_r$ are the variance of the translation in meters and the rotation in degrees, respectively.

\subsection{Vision Module}
\label{sec:overview:vision}

This module plays the role of coordinating the motions of multiple robots. All robots are registered to the stereo camera and move according to the task progress.

\subsubsection{Detection and Tracking for Continuous Tool Pose Estimation}
\label{sec:overview:tool}

In this work, we have applied a visual tracking and pose estimation scheme similar to~\cite{Zhang2017}. Bar-code markers with known geometrical characteristics were attached on each tool. A pentagonal adapter was used to ensure that the marker can be observed by the stereo cameras during manipulation (Fig.~\ref{fig:search}).

We used the marker detection algorithm in ArUco~\cite{Jurado2014}, combined with an optical flow based tracker in~\cite{Kalal2010}. It is worth mentioning at this point the forward-backward error identification component that we included for tracking.
We used the location of a marker in the previous frame to initialize a set of corner points. These points $\left\lbrace q_{i}\right\rbrace_{i=1}^{M}$ (belonging to the marker) were tracked ``forward'' from the previous to current frame, to obtain their estimated current locations $\left\lbrace q^{+}_{i}\right\rbrace_{i=1}^{M}$. In addition, ``backward'' tracking was also performed from the current to the previous frame, to obtain $\left\lbrace q^{-}_{i}\right\rbrace_{i=1}^{M}$.

The assumption used here is that if these points have been tracked accurately from the previous to the current frame, the backward tracking would return to the original locations of these points.
With this, we used the Euclidean measure to determine if a point estimate is valid, and we compared the measure with a threshold $\tau$ (defined as 1px). All the accepted points are treated as inliers, which are then used to estimate the 6 d.o.f pose of the marker using perspective-n-points \cite{Lepetit2009}. Hence, our pose estimation approach has the advantage of continuous pose estimation by combining visual detection and tracking which was applied to every marker on the adapter.

\subsubsection{Needle Detection}
\label{sec:overview:needle}
After being handed over twice in one stitching cycle, i.e., from Needle Driver A to B and then back to A, the needle may deviate from its initial pose relative to Needle Driver A. Although the change for each cycle may be small, it can accumulate to a large deviation from the initial needle pose, and cause task failure. Therefore, the robot motion needs to change adaptively according to the needle pose and to move the needle along its learned reference trajectory. To this end, the pose is estimated by performing a constrained 2D/3D rigid registration using features calculated in the image and a sparse representation of the 3D model of the needle, i.e., a set of 3D points along the needle shaft.

\begin{figure}
\centering
\subfloat[\scriptsize{}]{\includegraphics[height=2.5cm]{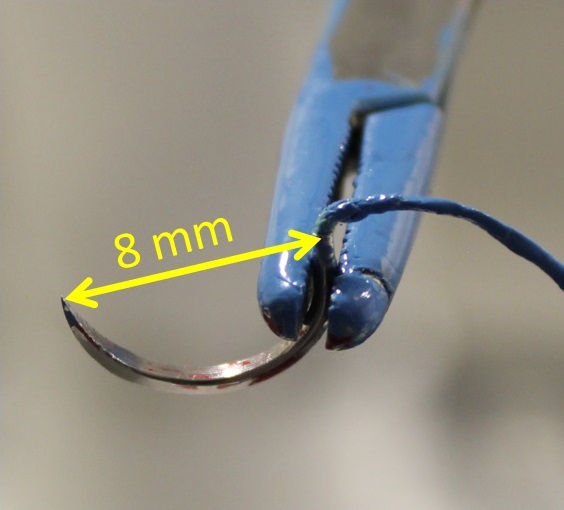}}
\hspace{5mm}
\subfloat[\scriptsize{}]{\includegraphics[height=2.5cm]{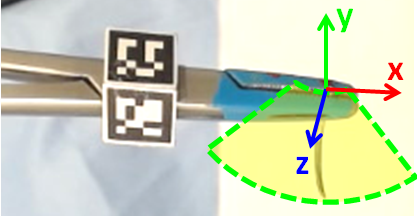}}
\caption{(a) Curved needle for sewing (b) A 2D illustration of the needle search space (yellow area). This search space is 4D and is restricted to $\pm$ 5 $mm$ for the x translation, $\pm$10 $degrees$ along x, $\pm$60 $degrees$ along y, $\pm$30 $degrees$ along z.}
\label{fig:search}
\end{figure}

This can be represented as a constrained 2D/3D rigid registration problem. For this purpose, the transformation that describes the pose of the needle is applied to its 3D model. The resulting 3D points are then projected onto the image using the camera's parameters\footnote{The camera's parameters are estimated during an offline calibration phase.}. Restricted by the jaws of the needle driver, the pose of the needle was represented in 4D: a translational movement along the jaws and the 3D rotation. For each plausible needle pose, the sum of the feature strength for the projected 3D model points is calculated (Fig.~\ref{fig:search}). Finally, the pose that is characterized by the highest overall feature score is regarded as the pose of the needle.
Due to the elongated shape of the needle, an image feature that has a strong response to lines and curvilinear objects~\cite{baert2003guide} was used. How the robot adapts its motion to the needle pose is explained in the next section.

\subsubsection{Visual Servoing}
\label{sec:overview:visualservoing}
A closed loop vision-based feedback system was deployed to guide the robot motions. In a multi-robot system, it is important to coordinate all the robots to work under the same frame of reference with the same pace.
Calibrating multiple robots is time-consuming, especially for tasks such as sewing or surgical tasks requiring high precision.
To this end, we applied a 3D visual servoing technique to ease the requirement of the accuracy of this calibration. With online visual feedback, the error of the robot reproducing the reference trajectory is independent of the calibration and the robot kinematic precision~\cite{hutchinson1996tutorial}.

In our sewing system, multiple reference frames are involved: camera ($c$), mandrel ($m$), stitching slots ($s$), needle ($n$), needle driver ($d$), robot base ($r$) and the robot end effector ($ee$).
Here, we denote $\tensor[^b]{\bullet}{_a}$ as the homogeneous matrix of the pose of the object $a$ in the frame of the object $b$.
Further, we use $\tensor[^b]{x}{_a}$ to denote a pose that changes along the robot motion, and $\tensor[^b]{H}{_a}$ to denote the relative pose between $a$ and $b$, which is a constant or independent of robot motion.

Prior to the task demonstration, the mandrel was registered to the end effector frame of Robot C, and each stitching slot was registered to the mandrel ($\tensor[^m]{H}{_s}$) according to the mandrel's design. Each personalized mandrel requires new registration.
The poses of the needle drivers (A, B) in the robots' (A,B) end effector frame were also computed.
All robots were registered to the frame of the camera ($\tensor[^c]{H}{_r}$). This was achieved by hand-eye calibration\footnote{From Matlab: https://uk.mathworks.com/matlabcentral/fileexchange/22422-absolute-orientation}.

For both the mandrel and the bimanual module, we adopted the ``look-and-move'' servoing method to control the robot movements. In this method, a robot is programmed to move to minimize the error between the current pose and the target pose of the observable objects, e.g. the markers on the mandrel and on the needle drivers. The location of the stitching slot used for demonstration ($s_0$) in the camera frame ($c$) was firstly registered to the camera by:

\begin{equation}
\setstretch{1.5}
{
\tensor[^c]x{_{s_0}} =
\tensor[^c]x{_m}\cdot
\tensor[^m]H{_{s_0}}
}
\end{equation}

Hence, for the mandrel to deliver the $i-th$ stitching slot to the same location, the error in pose was computed as:
\begin{equation}
\setstretch{1.5}
{
\tensor[^m]x{_{m_i}} =
\left(\tensor[^c]x{_{m}}\right)^{-1}\cdot
\tensor[^c]x{_{s_0}}\cdot\left(\tensor[^m]H{_{s_i}}\right)^{-1}
}
\end{equation}

This then can be transformed to find the error of Robot C end-effector by $\tensor[^{ee}]{H}{_m}$.  Hence we generate commands for Robot C to move the mandrel to the target pose. Note different mandrels will have different values of $\tensor[^m]H{_{s_i}}$.

The same principle was applied to control the bimanual sewing module. Taking Motion Primitive 1 as an example, the aim is to move the needle towards the stitching location and pierce the fabric. Hence the reference trajectory was represented as a series of needle poses in the frame of the stitching slot ($\tensor[^{s}]x{_{n}}$). The needle pose was transferred to the needle driver pose by:

\begin{equation}
\setstretch{1.5}
{
\tensor[^s]x{_{d}} =
\tensor[^s]x{_{n}}\cdot
\left(\tensor[^d]H{_{n}}\right)^{-1}
}
\end{equation}
where $\tensor[^d]H{_{n}}$ is the relative pose between the needle ($n$) and needle driver ($d$), detected over the task as explained in Section~\ref{sec:overview:needle}. With different needle poses, the robot will adapt its trajectory to ensure the $\tensor[^{s}]x{_{n}}$ remains the same, i.e. to produce the same stitch.

Similar to the mandrel module, the error between the current needle driver pose ($d$) and the desired needle driver pose ($d^*$) was computed as:

\begin{equation}
\setstretch{1.5}
{
\tensor[^d]x{_{d^*}} =
\left(\tensor[^c]x{_{d}}\right)^{-1}\cdot
\tensor[^c]x{_{s_i}}\cdot\tensor[^{s_0}]x{_{d^*}}
}
\end{equation}

This, again, can be converted to find the error of the robot end-effector and to control the robot to move to the desired location.

The frame rate of the camera is $20$ fps and hence is the control rate of visual servoing.
To increase the stability of visual servoing and to compensate for the latency of the cameras, a Double Rate Kalman Filter was applied to the estimation result of the mandrel and needle driver poses, as explained in our previous work~\cite{huang2017vision}.

\section{Experiments and results}
\label{sec:experiment}

\begin{figure*}
\captionsetup[subfigure]{width=2.5cm}
\centering{
\subfloat[\scriptsize{Trial 1}] {\includegraphics[height=5.5cm]{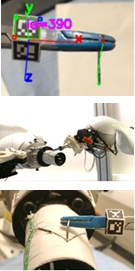}}
\hspace{1mm}
\subfloat[\scriptsize{Trial 2}] {\includegraphics[height=5.5cm]{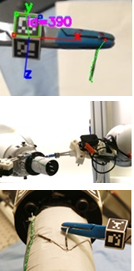}}
\hspace{1mm}
\subfloat[\scriptsize{Trial 3}] {\includegraphics[height=5.5cm]{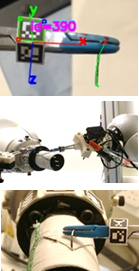}}
\hspace{1mm}
\subfloat[\scriptsize{Trial 4}] {\includegraphics[height=5.5cm]{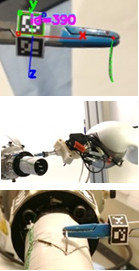}}
\hspace{1mm}
\subfloat[\scriptsize{Trial 5}] {\includegraphics[height=5.5cm]{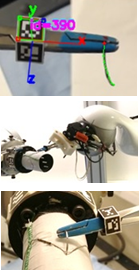}}
\hspace{1mm}
\subfloat[\scriptsize{Trial 6}] {\includegraphics[height=5.5cm]{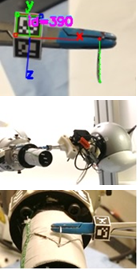}}

\caption{Experiment 1 results for the needle puncture task with six different initial needle positions. Top row: Results of needle 3D pose detection. The detected needle 3D poses are projected to 2D and represented by the green lines. The red dots mark the end of the detected needle. The pose of the needle driver is detected by the marker, denoted by the frame. Middle row: The robot adaptation of the needle poses during the puncture task execution. Bottom row: At the end of the task, the needle punctured the fabric.}
\label{fig:needledetection}
}\end{figure*}

\begin{table}[]
\fontsize{8}{12}\selectfont
\centering
\caption{Qualitative results of needle puncture task. Trials 1, 3 and 4 have no observable errors (denoted by -).}
\label{tab:needledetection}
\begin{tabular}{cccccc}
\hline
\rowcolor[HTML]{C0C0C0}
\cellcolor[HTML]{C0C0C0} & $\theta_x$ & $\theta_y$ & $\theta_z$ & X & Error \\
\rowcolor[HTML]{C0C0C0}
\multirow{-2}{*}{\cellcolor[HTML]{C0C0C0}\begin{tabular}[c]{@{}c@{}}Needle\\ pose\end{tabular}} & (degrees) & \multicolumn{1}{l}{\cellcolor[HTML]{C0C0C0}(degrees)} & \multicolumn{1}{l}{\cellcolor[HTML]{C0C0C0}(degrees)} & (mm) & \multicolumn{1}{l}{\cellcolor[HTML]{C0C0C0}(mm)} \\ \hline \hline
1 & -1.00 & 0 & 8.51 & -1 & - \\ \hline
2 & 0.41 & -9.39 & -0.12 & 0 & 1.63 \\ \hline
3 & 0.12 & -0.99 & 1.50 & -2 & - \\ \hline
4 & 1.17 & 6.49 & 10.0 & 1 & - \\ \hline
5 & -2.88 & 21.00 & 7.43 & -2 & 0.5 \\ \hline
6 & -2.00 & 6.00 & 13.53 & -3 & 0.8 \\ \hline
\end{tabular}
\end{table}

Two tasks were implemented to evaluate the performance (accuracy and the robustness) of the proposed system: 1) a needle puncture task and 2) the autonomous sewing of personalized stent grafts.

\subsection{Needle Puncture Task}
The needle detection and the visual servoing algorithm were evaluated with a puncture task: teaching the robot to puncture a fixed point on a fabric with a needle. This motion was demonstrated by a user using a needle driver to grip and move the needle. At the beginning of the motion, the needle was placed away from the fabric and the needle pose was estimated by the algorithm presented in Section~\ref{sec:overview:needle}. The user moved the needle to approach the fabric and to pierce the fabric at a given location. The motion trajectory of the needle driver was recorded by the vision system (resolution: 640$\times$480). The robot then learned to reproduce the trajectory and puncture the fabric at the same location. The locations of the puncture points in different trials were recorded. These demonstrations were performed using a 1/2 circular needle with 8$mm$ diameter (Fig.~\ref{fig:search}), of which the model has 10 evenly distributed points along the arc.

Six experiments were conducted and at each trial, the needle pose varied as shown in Fig.~\ref{fig:needledetection}. After needle pose estimation, the robot adapted its motion trajectory to deliver the needle to pierce at the same point. The motion was reproduced at a third of the speed of human demonstration. The error of each experiment was computed as the distance of the repeated puncture points with respect to the original puncture point. This error reflects the overall accuracy of the entire vision module, including the needle detection algorithm, tool tracking, and the visual servoing method.

As shown in Table~\ref{tab:needledetection}, in half of the trials the robot pierced at the same point as demonstrated. The average error across the six trials was 0.48 $mm$. Trial 2 resulted in a larger error (1.63 $mm$) as during the motion, the robot arm approached its joint limit and hence could only reach the adjacent point. This result shows that the accuracy of the vision module allows for high precision sewing for the stent grafts.
The accuracy can be improved significantly by using a higher resolution stereo vision system~\cite{perez2016robot}.

\begin{figure}
\centering
{
\subfloat[\scriptsize{A}] {\includegraphics[height=3cm]{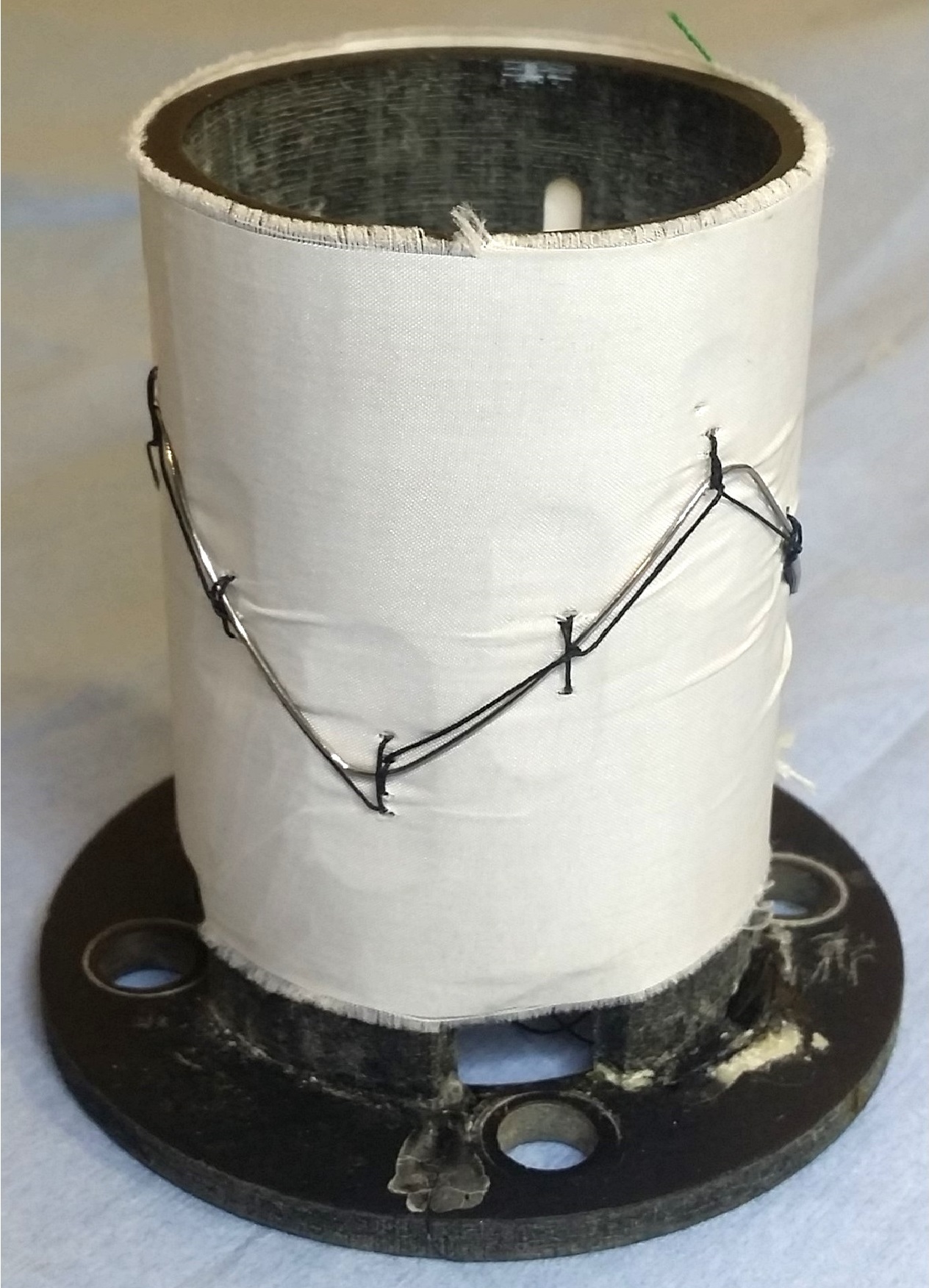}}
\subfloat[\scriptsize{B}] {\includegraphics[height=3cm]{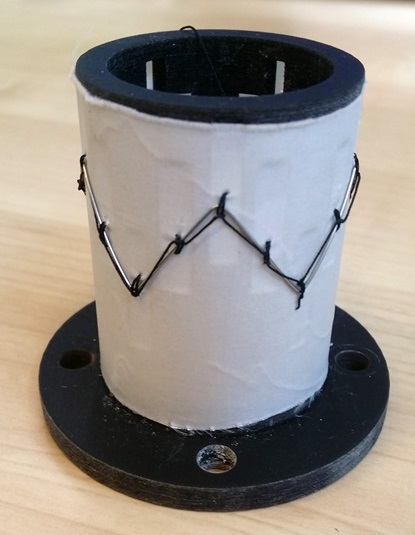}}
\subfloat[\scriptsize{C}] {\includegraphics[height=3cm]{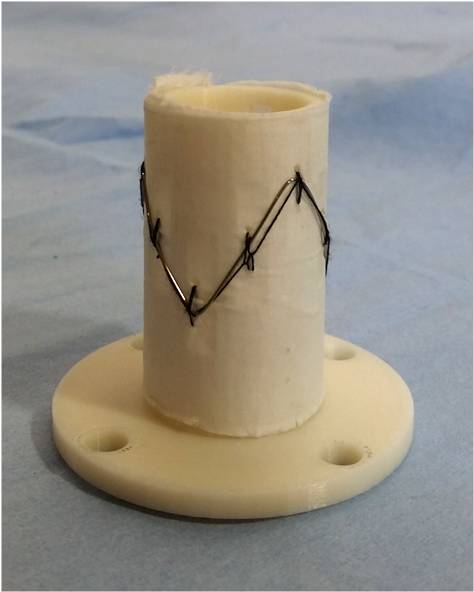}}
\subfloat[\scriptsize{D}] {\includegraphics[height=3cm]{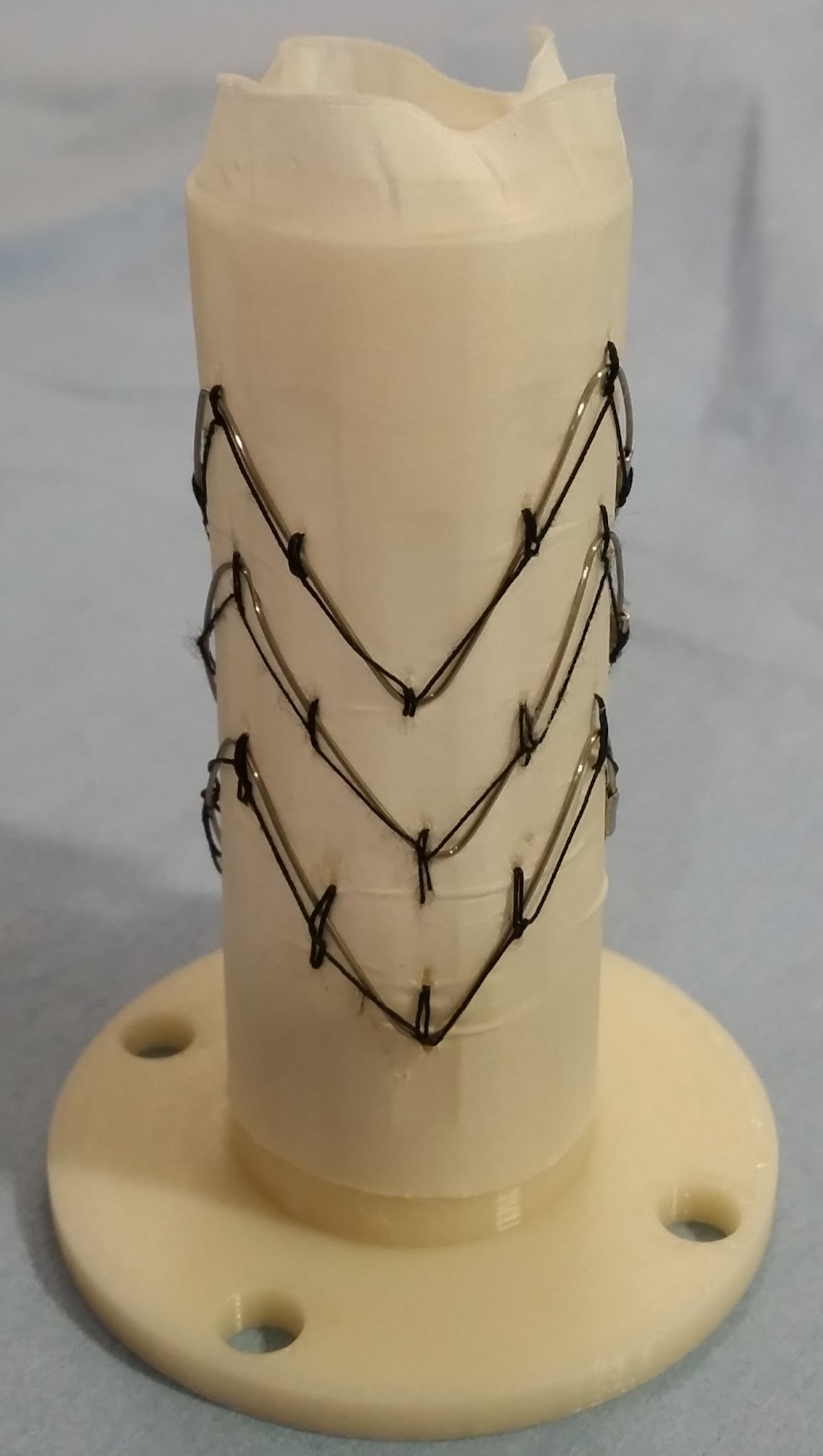}}
\caption{Stent grafts in different designs. From left to right, the outer diameters are: 4.4cm, 4cm, 3cm, 3cm. }
\label{fig:stents}}
\end{figure}

\begin{figure*}
\captionsetup[subfigure]{width=2.5cm}
\centering{
\subfloat[\scriptsize{Primitive 1 (in frame of mandrel)}] {\includegraphics[width=3.5cm]{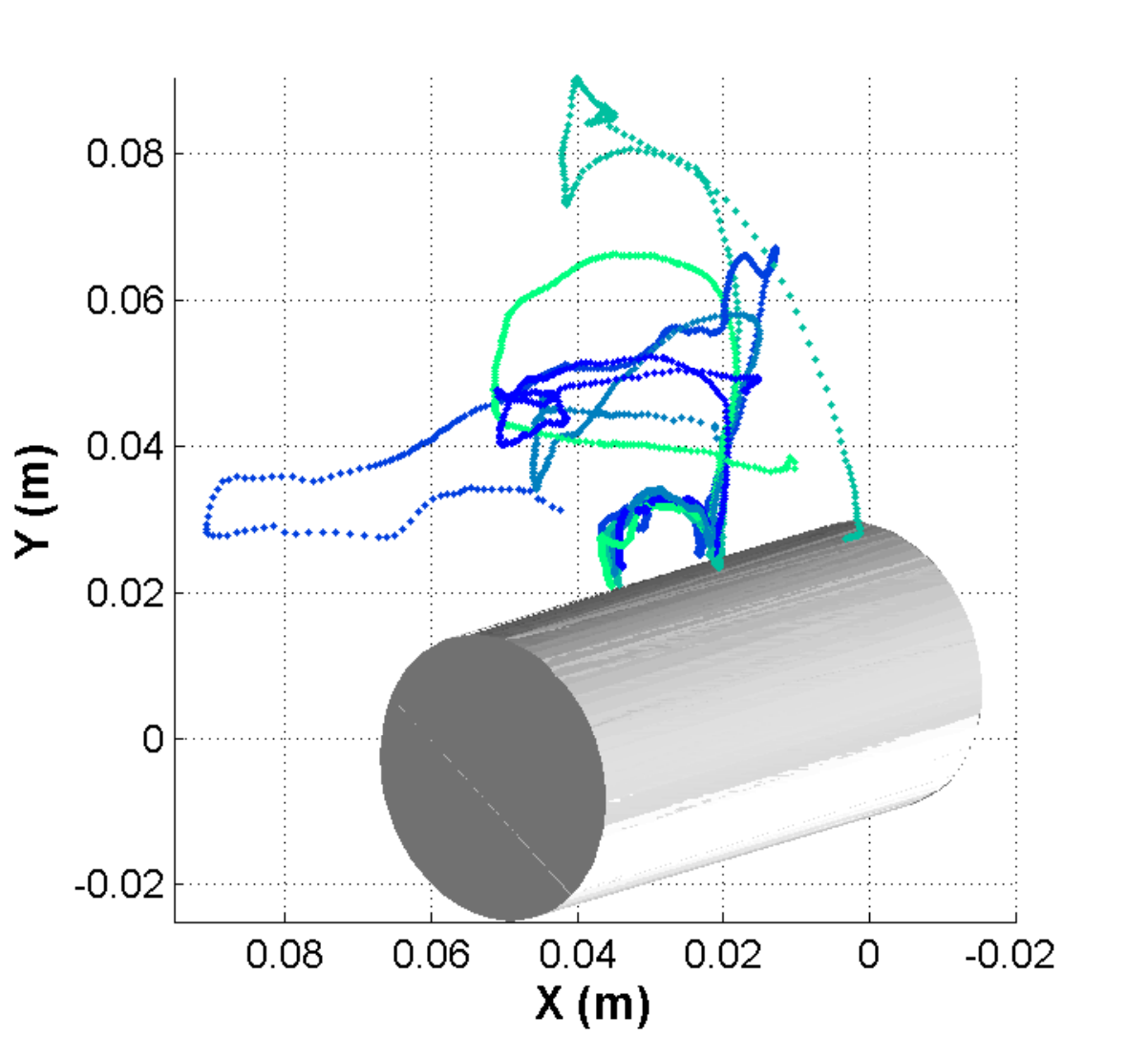}}
\subfloat[\scriptsize{Primitive 2 (in frame of mandrel)}] {\includegraphics[width=3.5cm]{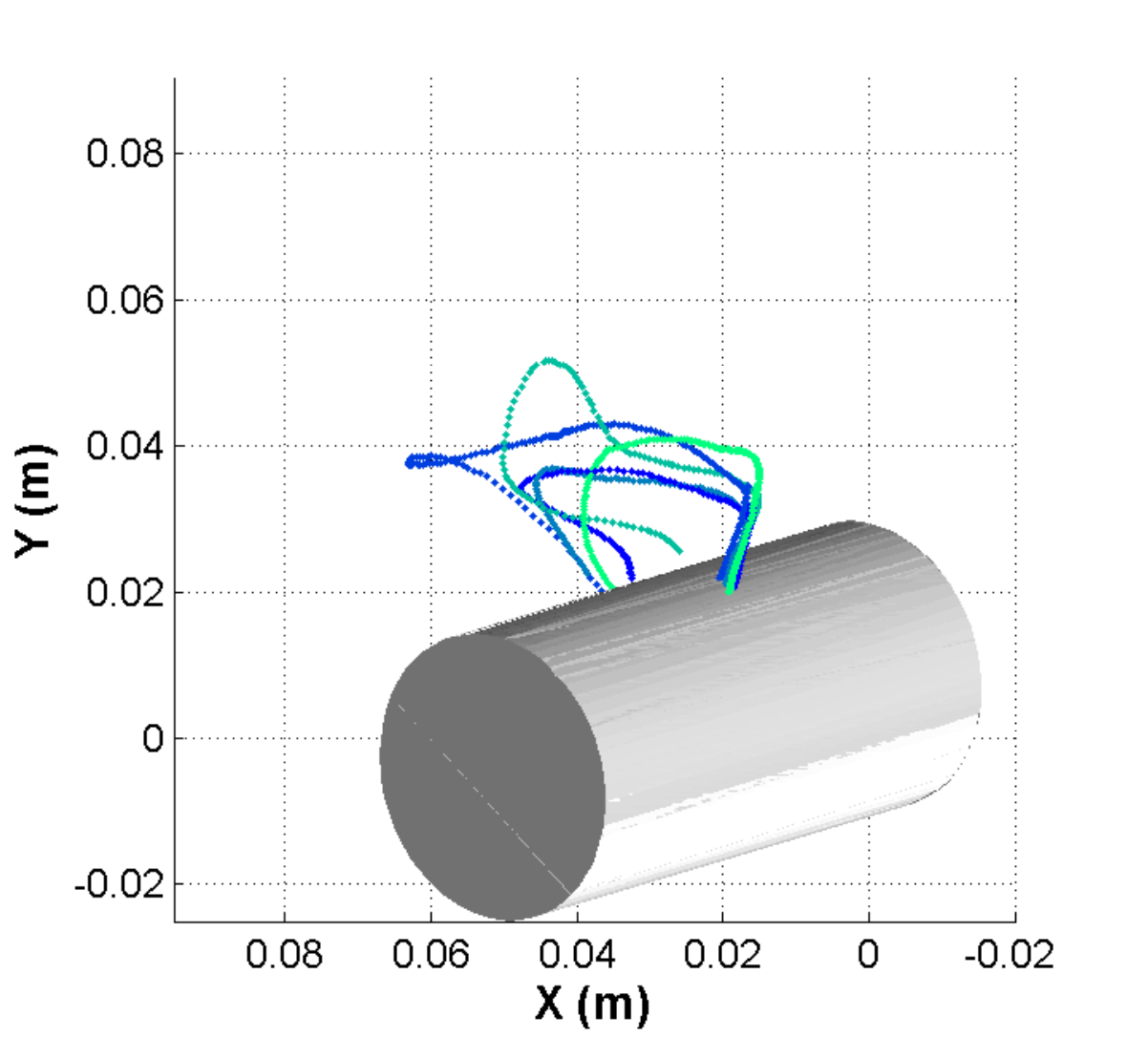}}
\subfloat[\scriptsize{Primitive 3 (in frame of mandrel)}] {\includegraphics[width=3.5cm]{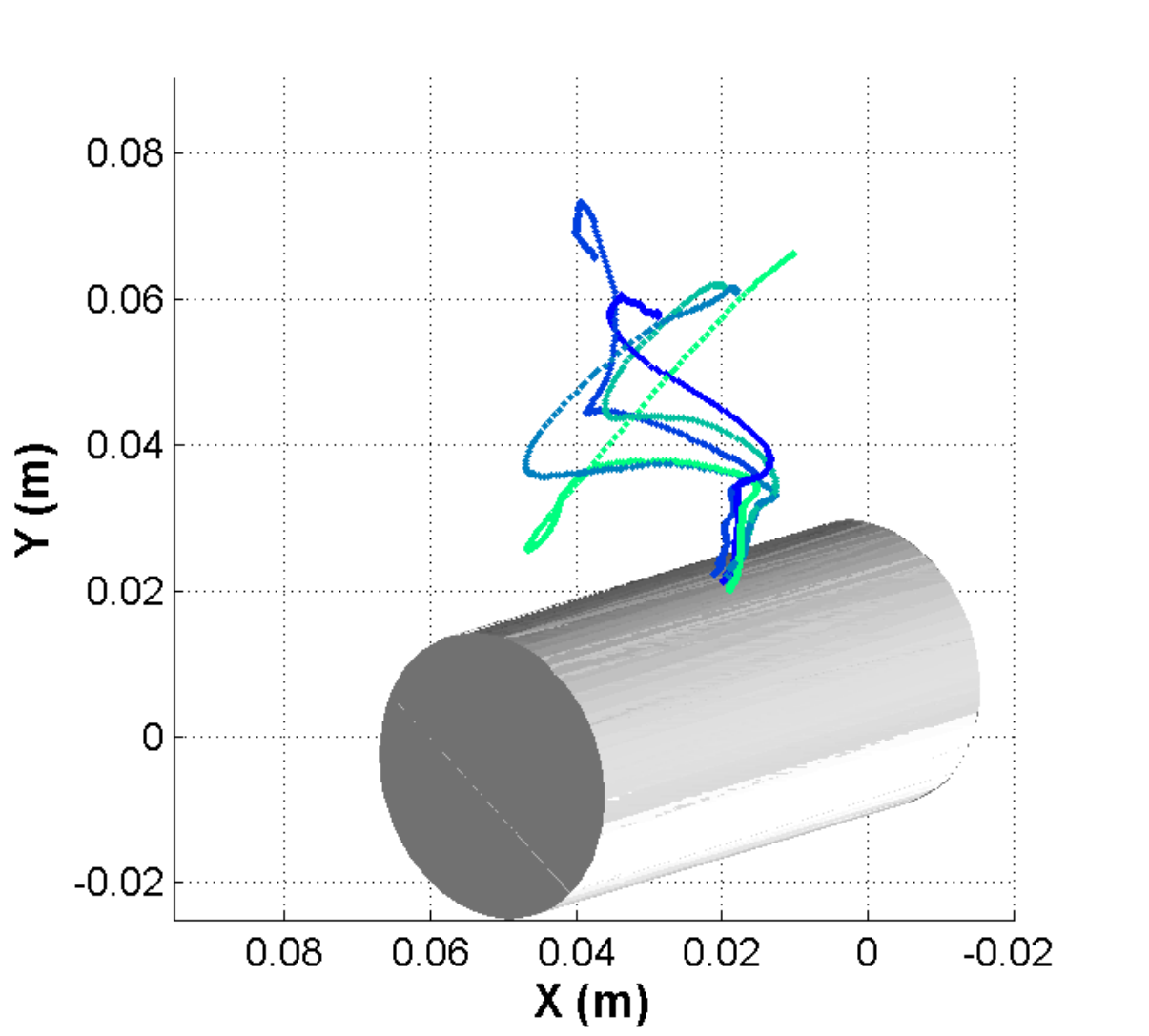}}
\subfloat[\scriptsize{Primitive 4 (in frame of needle)}] {\includegraphics[width=3.5cm]{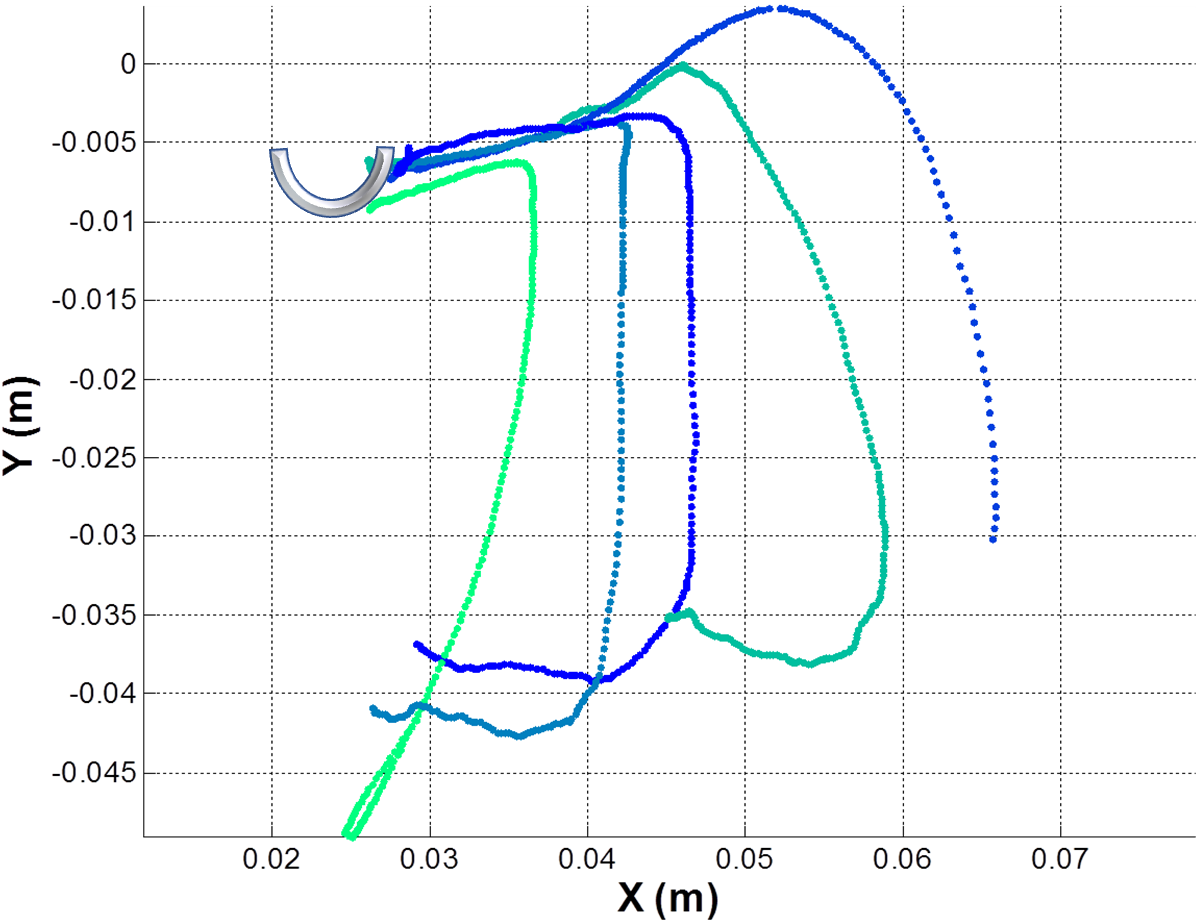}}
\subfloat[\scriptsize{Primitive 5 (in frame of needle's initial pose)}] {\includegraphics[width=3.5cm]{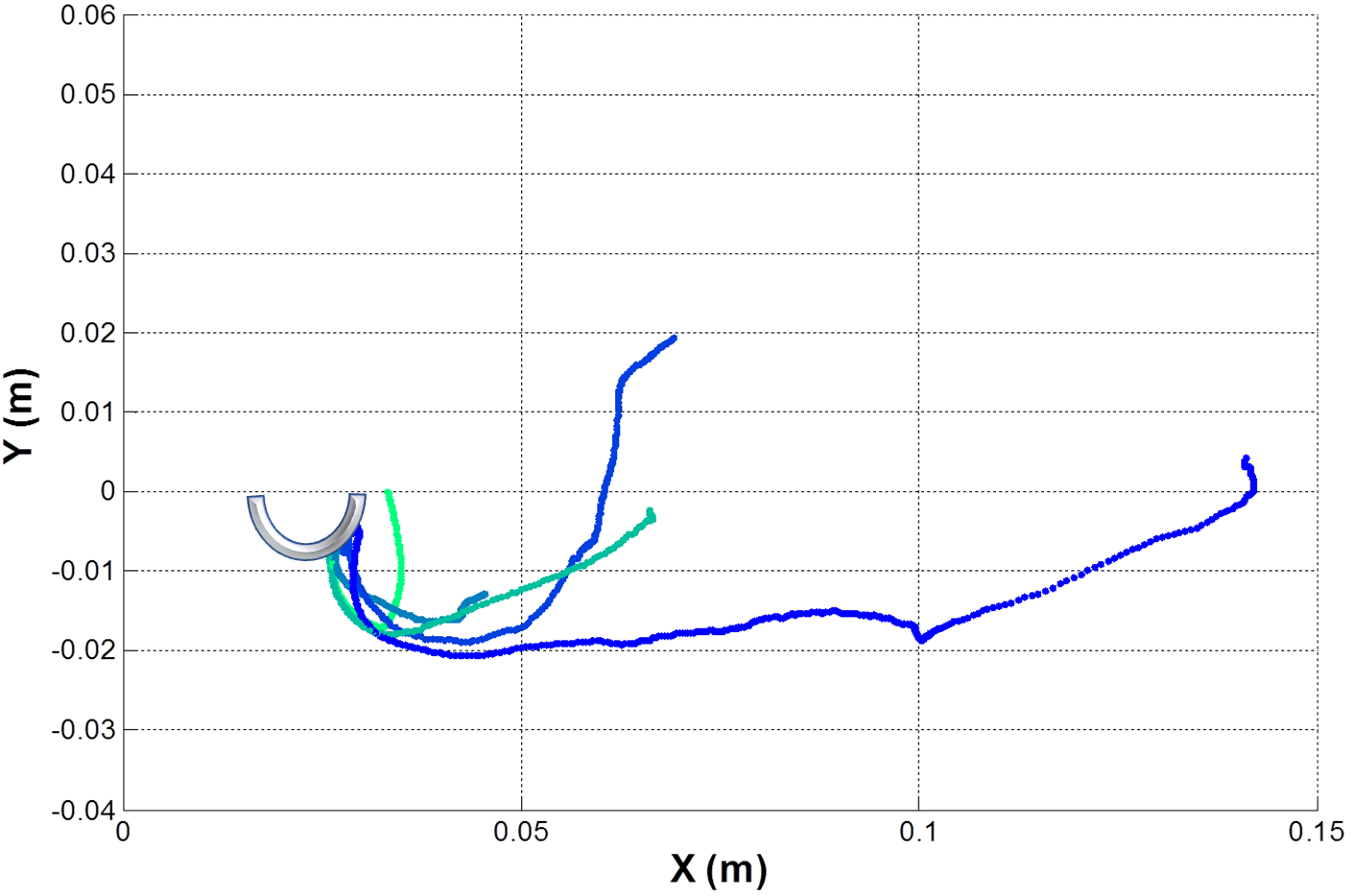}}

\subfloat[\scriptsize{Needle Driver A motion (15 seconds)}] {\includegraphics[width=3.5cm]{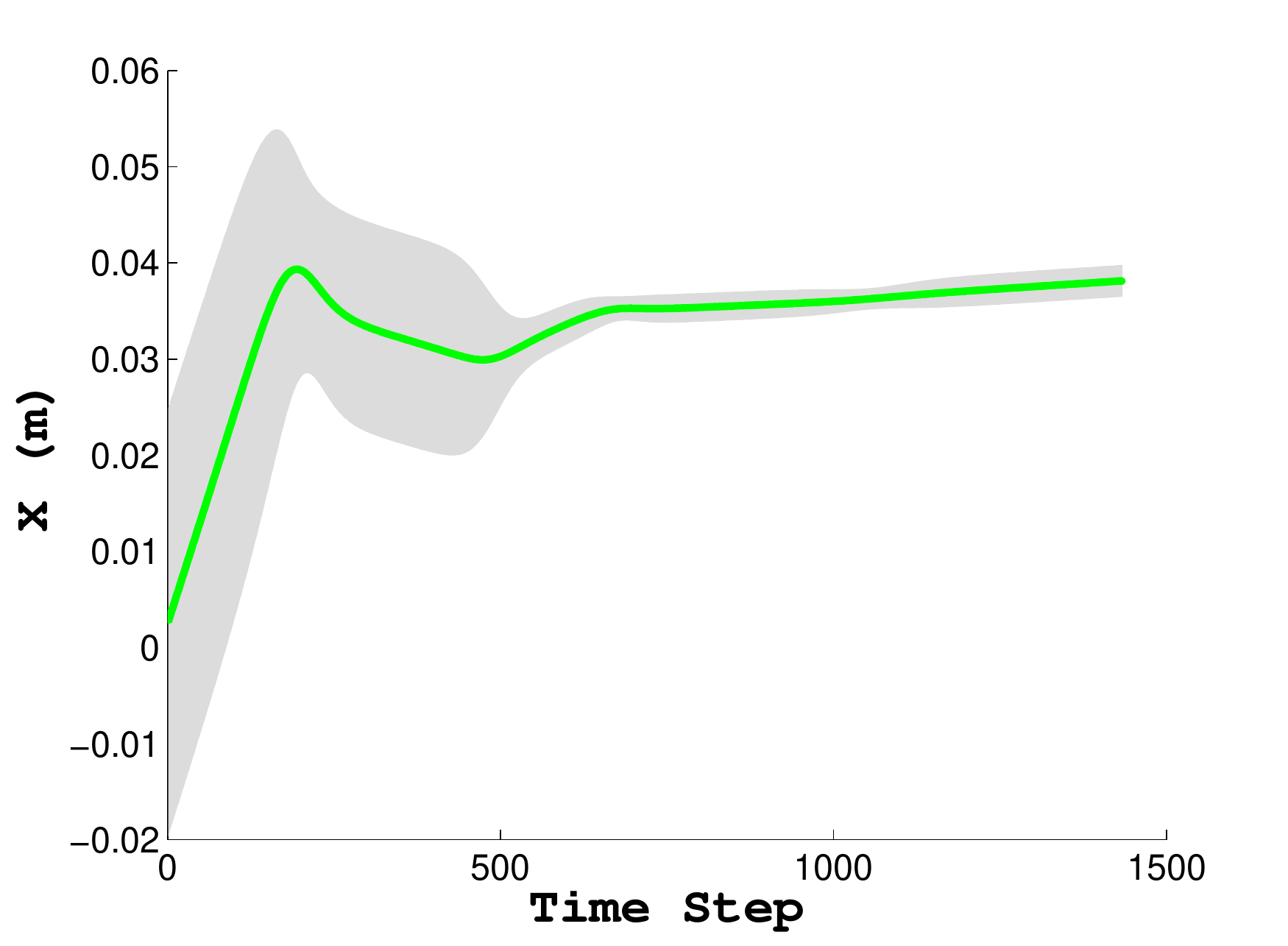}}
\subfloat[\scriptsize{Needle Driver B motion (6 seconds)}] {\includegraphics[width=3.5cm]{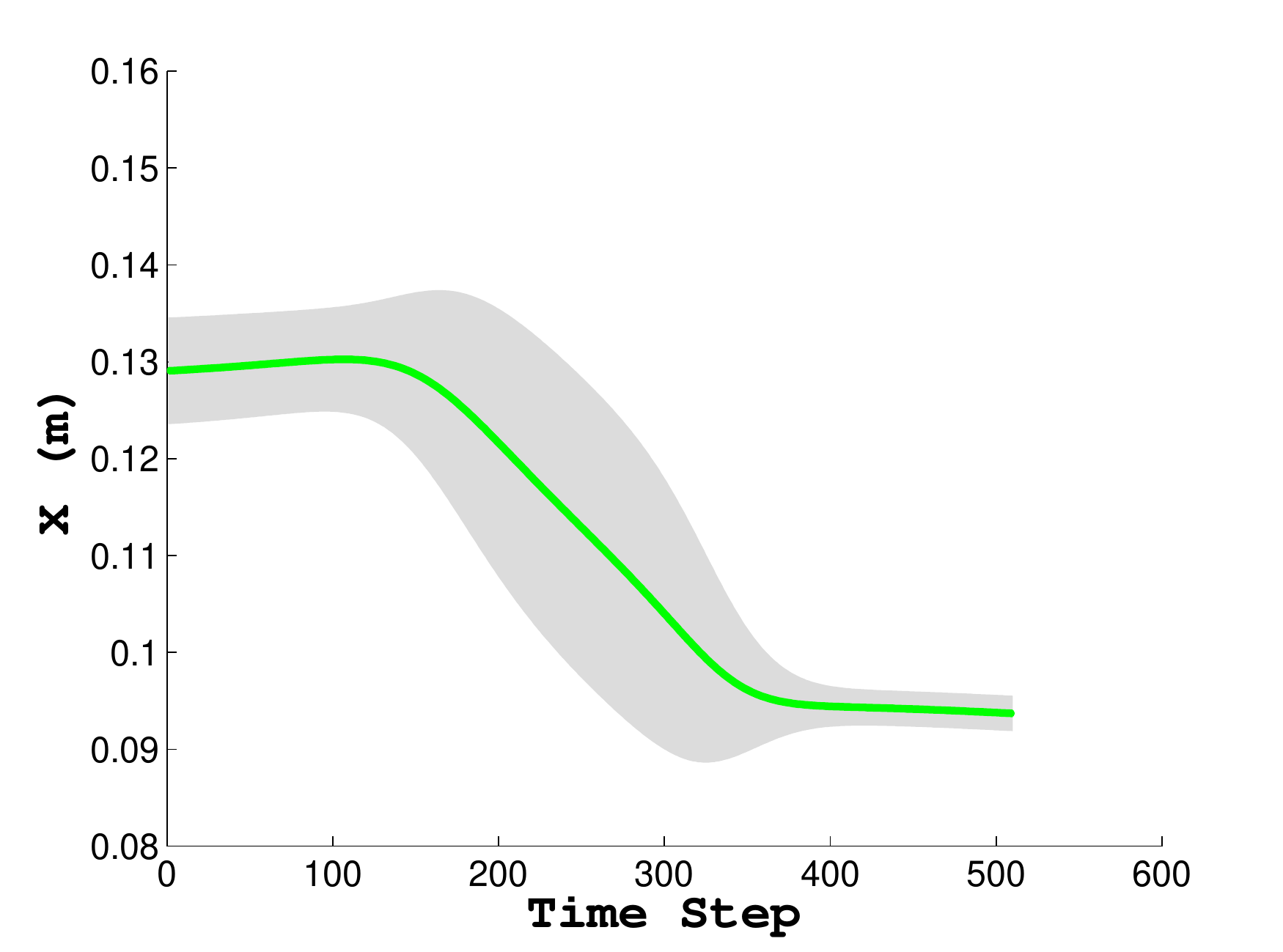}}
\subfloat[\scriptsize{Needle Driver B motion (8 seconds)}] {\includegraphics[width=3.5cm]{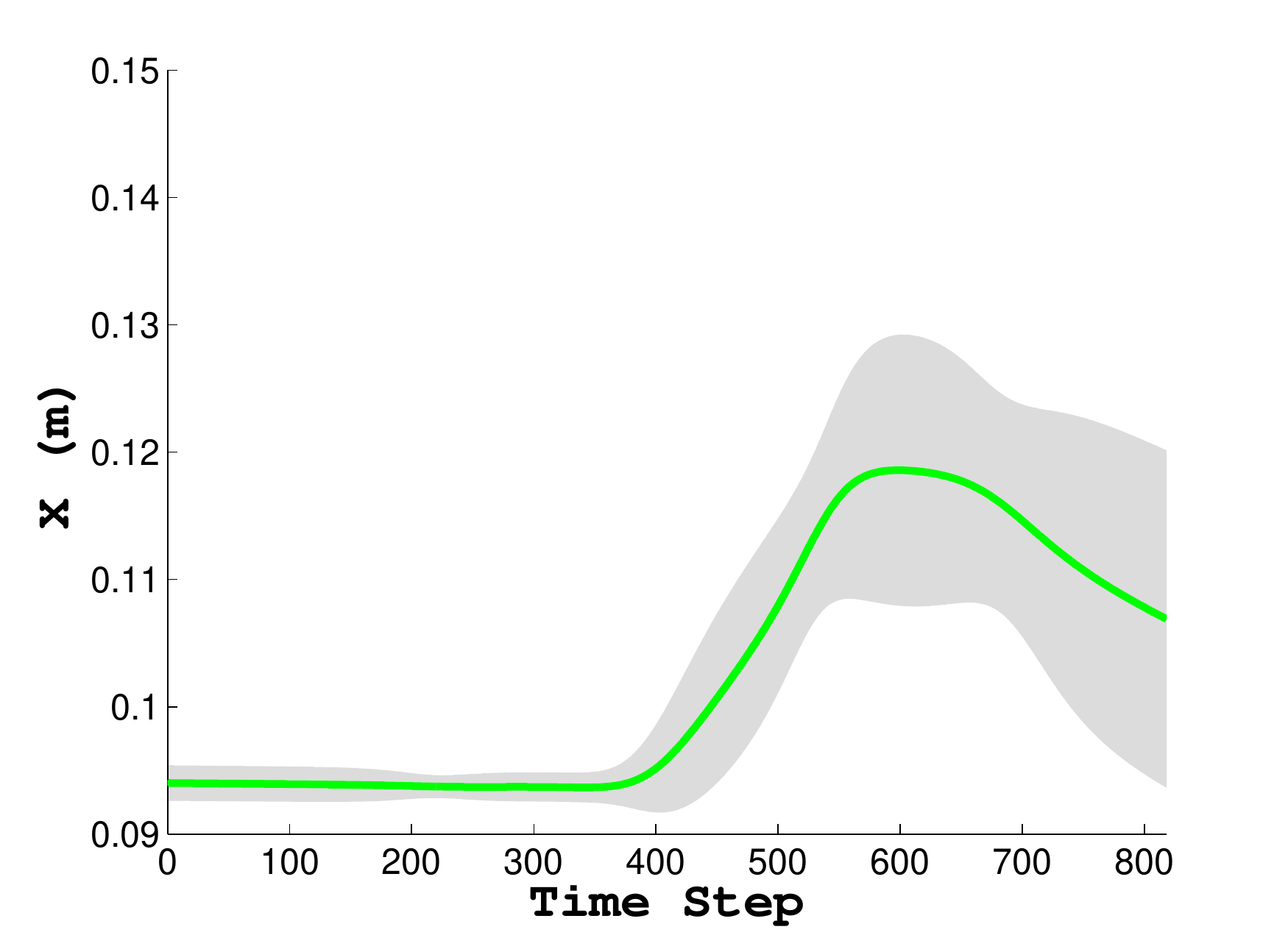}}
\subfloat[\scriptsize{Needle Driver A motion (4 seconds)}] {\includegraphics[width=3.5cm]{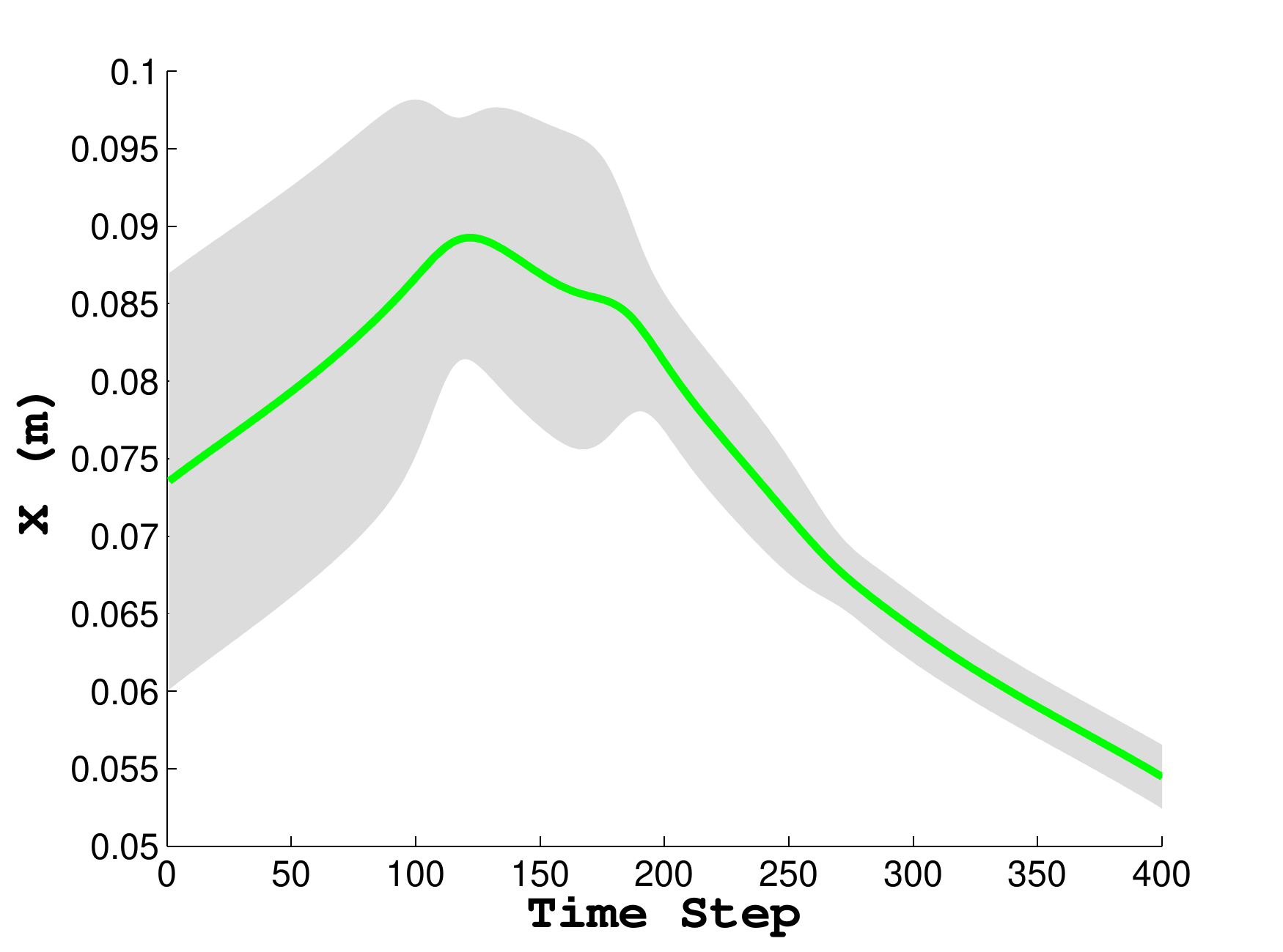}}
\subfloat[\scriptsize{Needle Driver A motion (8 seconds)}] {\includegraphics[width=3.5cm]{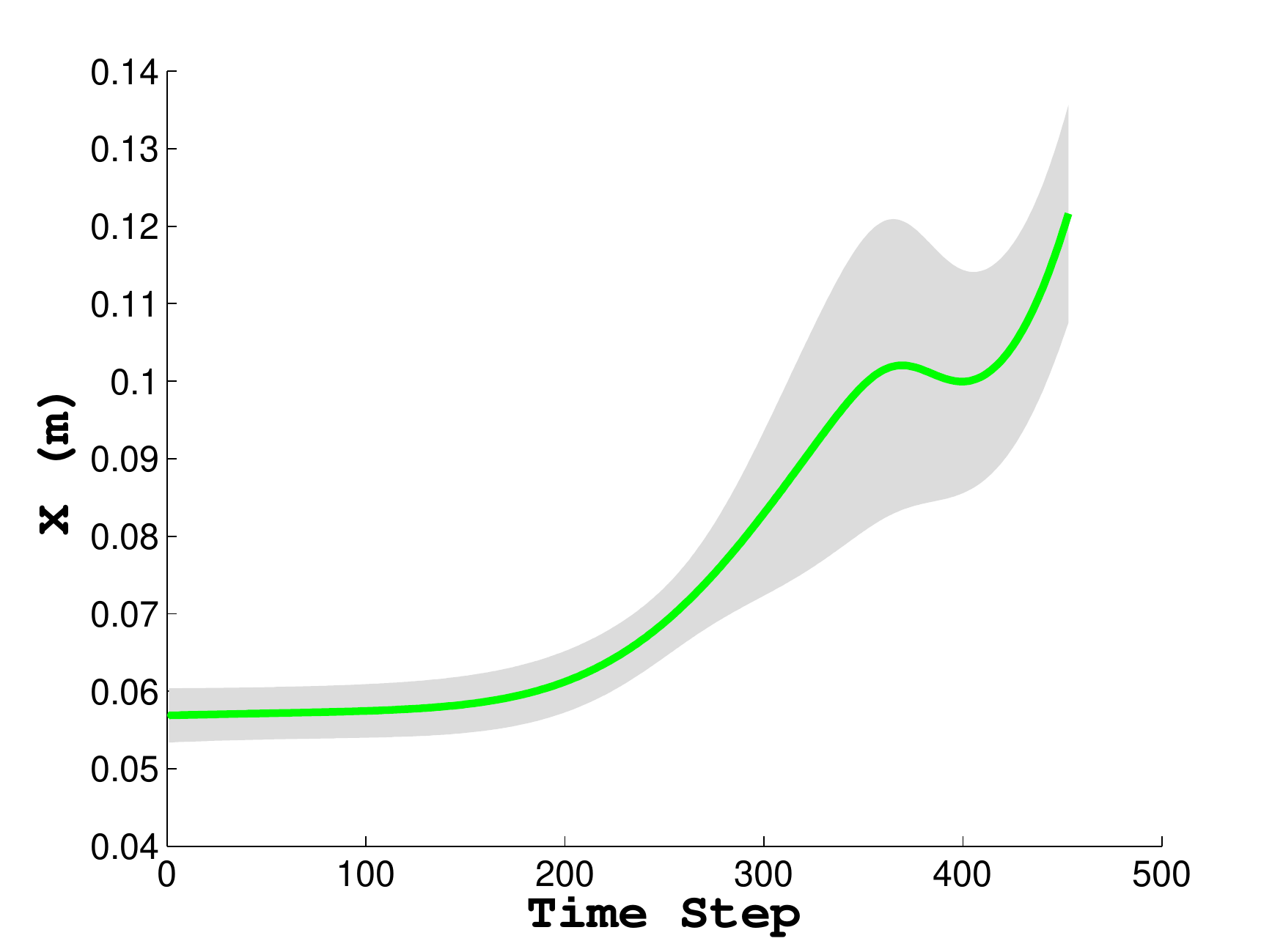}}

\caption{The needle drivers trajectories from human demonstrations and the learned reference trajectories of each motion primitive. (a)-(e): Five human demonstrations of bimanual sewing. Different colors represent different trials. The grey cylinder in (a)(b)(c) represents the mandrel. The grey arc in (d)(e) represents the needle. (f)-(j): 2D projection of the motion primitives on the x-axis. Green lines represent the reference trajectories, and the grey area represent the corresponding variances. }
\label{fig:demo}
}
\end{figure*}

\begin{figure*}
\centering
{
\subfloat[\scriptsize{Primitive 1}] {\includegraphics[width=3.4cm]{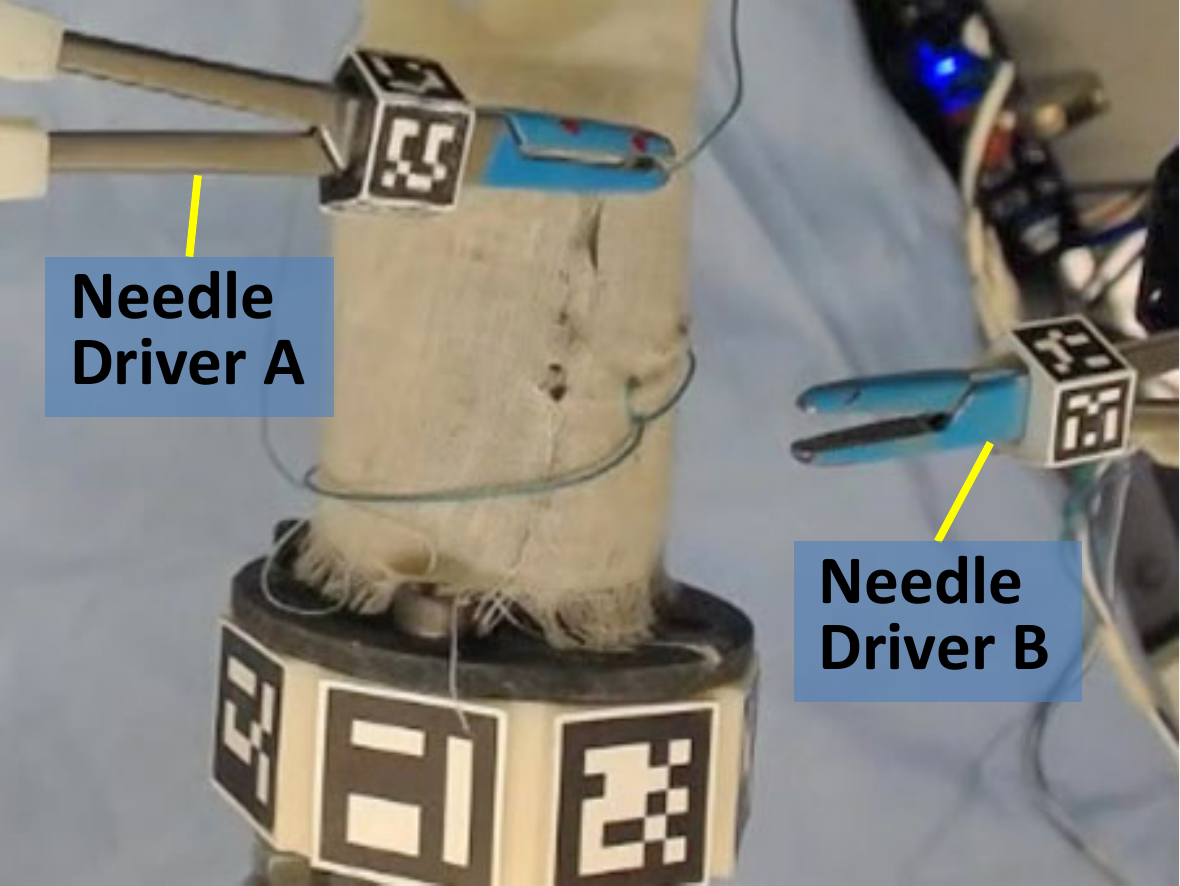}}
\hspace{1mm}
\subfloat[\scriptsize{Primitive 2}] {\includegraphics[width=3.4cm]{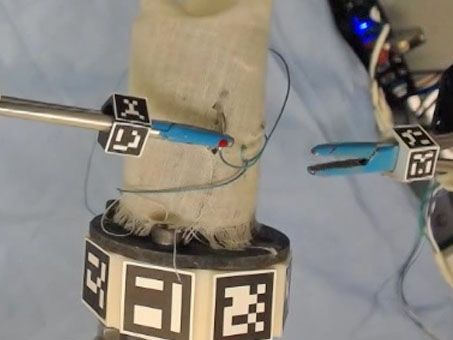}}
\hspace{1mm}
\subfloat[\scriptsize{Primitive 3}] {\includegraphics[width=3.4cm]{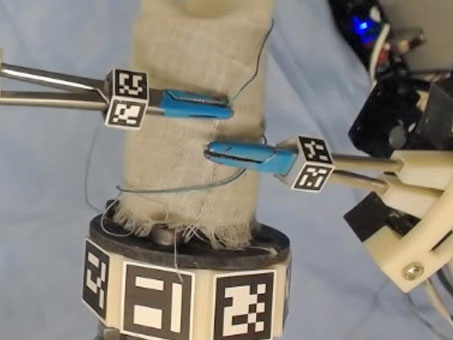}}
\hspace{1mm}
\subfloat[\scriptsize{Primitive 4}] {\includegraphics[width=3.4cm]{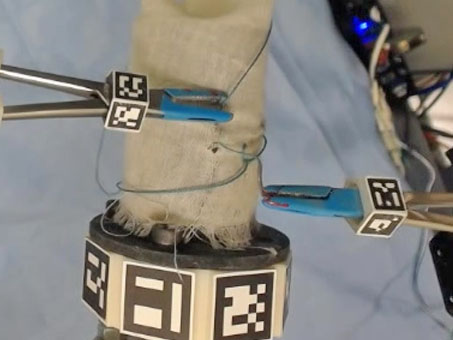}}
\hspace{1mm}
\subfloat[\scriptsize{Primitive 5}] {\includegraphics[width=3.4cm]{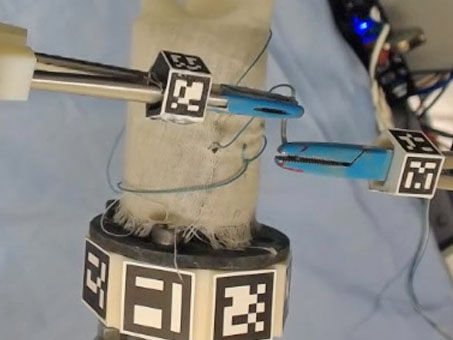}}

\subfloat[\scriptsize{Primitive 1}] {\includegraphics[width=3.4cm]{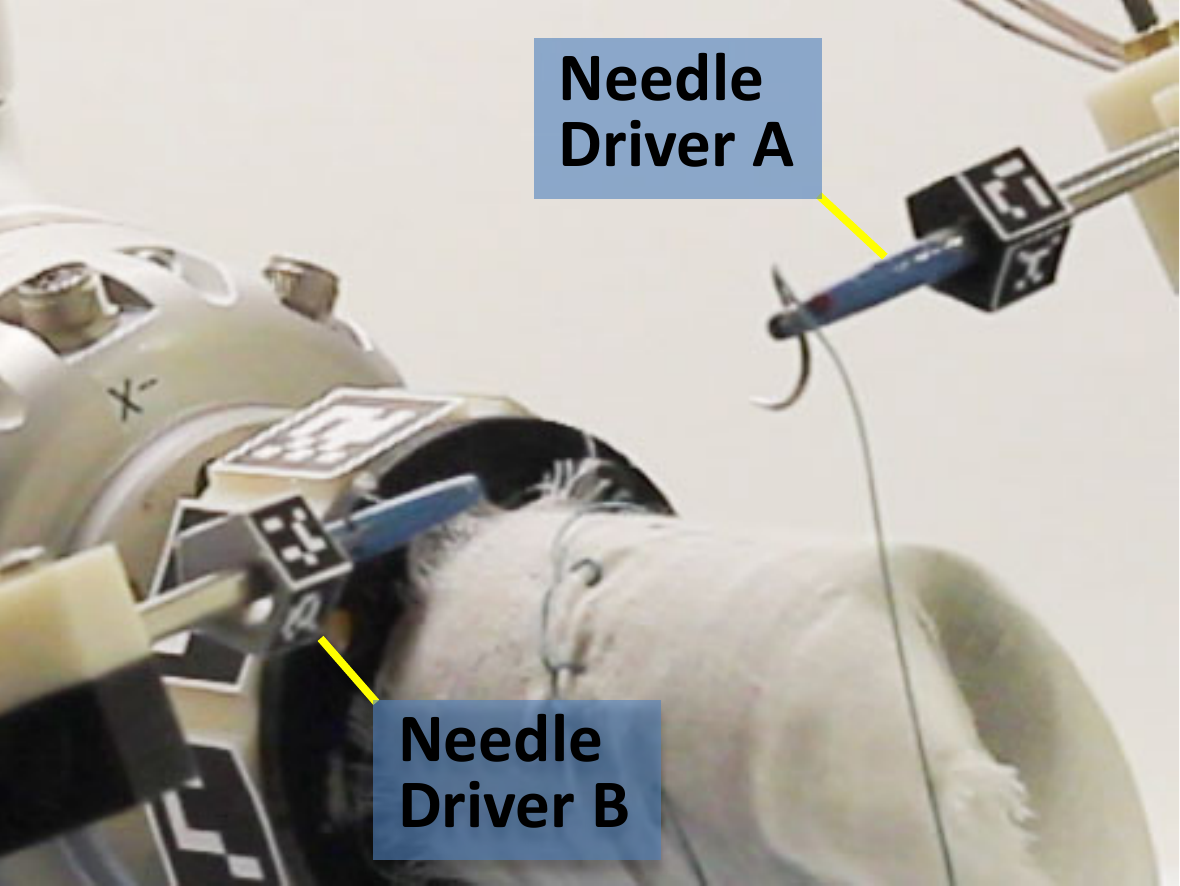}}
\hspace{1mm}
\subfloat[\scriptsize{Primitive 2}] {\includegraphics[width=3.4cm]{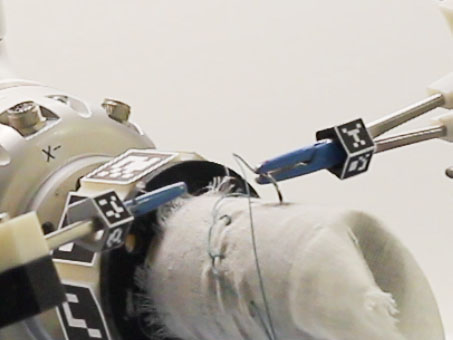}}
\hspace{1mm}
\subfloat[\scriptsize{Primitive 3}] {\includegraphics[width=3.4cm]{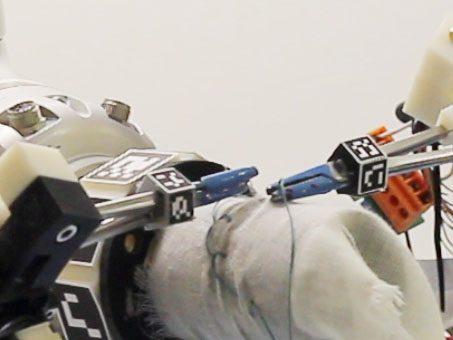}}
\hspace{1mm}
\subfloat[\scriptsize{Primitive 4}] {\includegraphics[width=3.4cm]{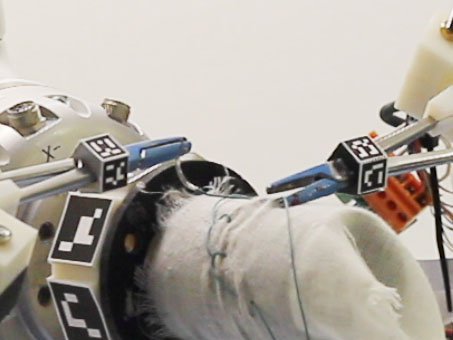}}
\hspace{1mm}
\subfloat[\scriptsize{Primitive 5}] {\includegraphics[width=3.4cm]{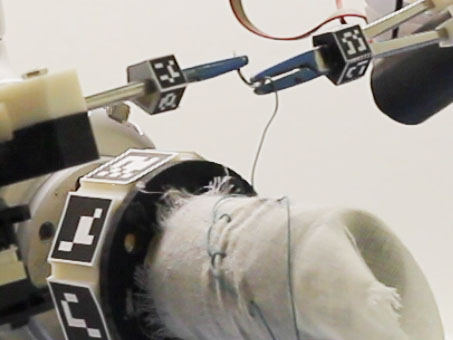}}
\caption{Key frames of bimanual sewing in each motion primitive. (a)-(e) The view from the top camera, used for visual servoing. (f)-(j) The corresponding views from the side.}
\label{fig:bimanual}}
\end{figure*}

\begin{figure*}
\centering
{
{\includegraphics[width=18cm]{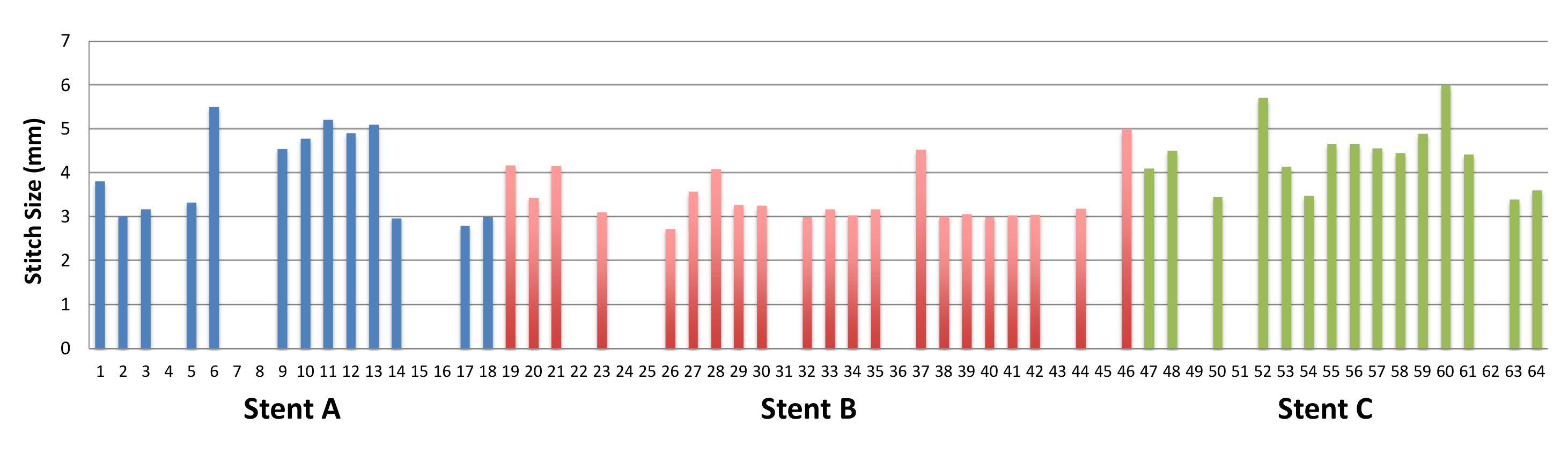}}
\caption{Experiment results of sewing stent graft A (blue:1-18), B (red: 19-46) and C (green: 47-64). In total 64 trials have been taken and the success rate is 77$\%$ with mean stitch size 3.93 $mm$ and variance 0.77 $mm$. Trials with no stitch size denote failed stitches. }
\label{fig:columnABC}}
\vspace{-3mm}
\end{figure*}

\begin{figure*}
\centering
{
{\includegraphics[width=18cm]{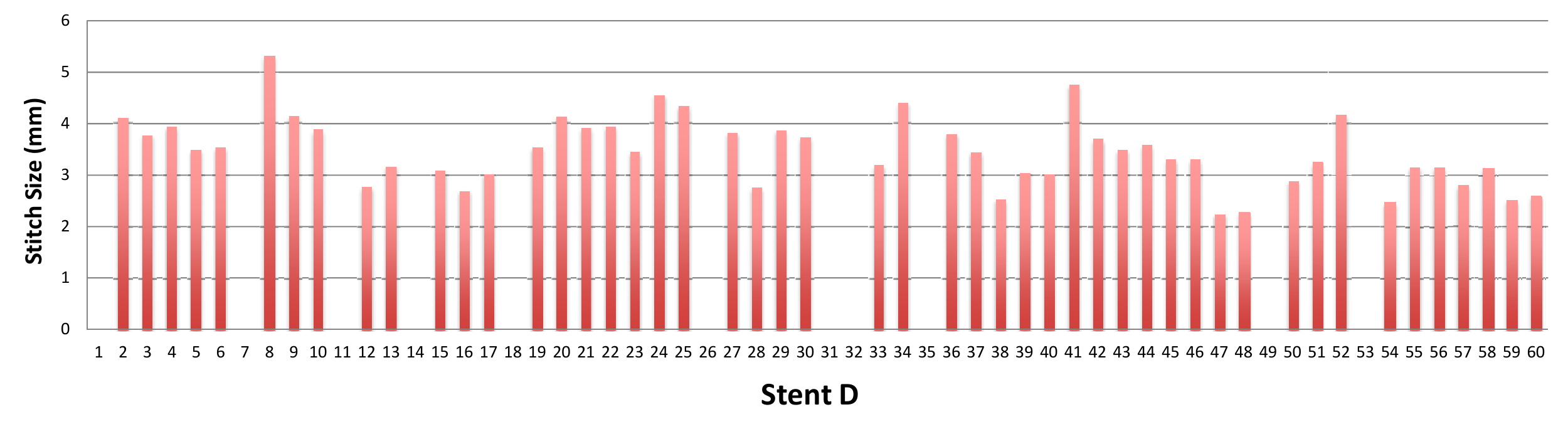}}
\caption{Experiment results of sewing stent graft D. In total 60 trials have been taken and the success rate is 82$\%$ with mean stitch size 3.46 $mm$ and variance 0.44 $mm$. Trials with no stitch size denote failed stitches.}
\label{fig:columnD}}
\vspace{-3mm}
\end{figure*}

\subsection {Autonomous Sewing of Personalized Stent Grafts}

To evaluate the performance of the proposed multi-robot sewing system, four experiments were conducted. These stents were different in designs and sizes (Fig.~\ref{fig:stents}). These sizes were chosen from the personalized sizes available from current stent graft manufacturers. The fabric was Dacron and the stent rings were medical stainless steel.

According to the design of these stents, their corresponding mandrels were 3D printed. On each mandrel, 10 $\times$ 2$mm$ stitching slots were created\footnote{Currently the smallest diameter of the mandrel we used was 3 $cm$ as limited by the rigidity of the 3D printing material. This can be made smaller if different printing material is used. In any case, most stent grafts for AAA have diameter larger than 3 $cm$.}, located at the peaks of the stents and at the middle point between the peaks. Before sewing, the fabric tube and the stent were manually loaded onto the mandrel and the whole device was mounted on Robot C.

The robotic sewing system is shown in Fig.~\ref{fig:setup}. All three 7 d.o.f robots were registered to the vision module mounted on top of the workspace by hand eye calibration. Bimanual sewing was demonstrated to the system five time on the same slot of Stent A by the user. The location of this slot was recorded. The average demonstrated stitch size was 4.10 $mm$. Before each stitch, the needle pose was detected such that the needle pathways can be computed by the needle driver trajectory.

The human demonstration motion primitives are shown in Fig.~\ref{fig:demo}. Note we have omitted in the figure the trajectories of Needle Driver A in Primitives 2,3 and the Needle Driver B in Primitives 1,4,5 as they are nearly static. As presented at the bottom row, the variance of each motion primitive varies across different stages.

For example, in Primitive 1 the motion variance of Needle Driver A is large at the beginning, i.e., approaching the fabric, and rapidly reduces, i.e. piercing fabric. This suggests that piercing motions were highly similar among all demonstrations and thus required to be followed precisely. For the same reason, in Primitives 2 and 3, the variance of the Needle Driver B motion is large when approaching or leaving the fabric, and small when piercing out the needle. Primitives 4 and 5 are shown in the needle frame. According to the variance, Needle Driver A gripped the same place on the needle for every demonstration.

In short, for the interactive parts of the task, i.e. needle piercing in/out of the fabric and needle handing over, the variance of the motion is small and hence we slow down the robot for these parts to ensure precision. For the other parts we increase the velocity of the robot to maintain the speed of sewing.

The learned reference trajectories were registered to the mandrel's frame via its stitching slots.
As mentioned in Section~\ref{sec:overview:vision}, the mandrel's pose was detected according to the fixed markers on the adapter. At the end of each stitch cycle, Robot C would rotate and translate the mandrel to the desired location to allow for easy access to the next stitching slot. The motion of Robot C was planned offline for each personalised stent graft.

Before the bimanual sewing module performed each stitch, the reference trajectory was adapted according to the detected needle pose. After finishing a stitch, Robot A was programmed manually to tighten the stitch. This pulling motion was stopped when the force sensor reading went above a predefined threshold value. When a stitch failed, the system was restored to the initial state of the stitching cycle and the sewing was restarted.

In total 124 stitches were made with an overall success rate of 79$\%$ and the average stitch size was 3.60 $mm$ with a variance of 0.89$mm$. The size of each stitch is reported in Fig.~\ref{fig:columnABC} and Fig.~\ref{fig:columnD}.
We identified four possible causes of a failed stitch: 1) needle handling failure (stent and trial A4, A7, A8, B36, B43, B45, D7, D14, D35, D49); 2) stitch-stent missing failure (A15, A16, B31, C49, C51, D11, D31, D32); 3) needle-stent touching failure (B24, B25, D26, D53); 4) needle-thread entangling failure (B22, C51, D1, D18). The first three causes were mainly due to errors in the needle detection and visual servoing, while the fourth was caused by the lack of thread shape control.

This experiment shows the proposed system can sew different stent graft designs effectively. The success rate of 79$\%$ can be improved by using higher resolution cameras. Advanced design of the needle driver and mandrel will also improve the system performance. For example, currently the maximum opening of a needle driver is 4 $mm$. The motorization design can be modified to allow the needle drivers to open wider and hence have higher tolerance of the position error of the needle.

The variance of the stitch size is mainly caused by the slack of the fabric at the stitching slots. A small deformation of the fabric can result in a large difference in the stitch size.
This can be improved by using a collapsible mandrel.
A mandrel with slightly larger diameter will bind the fabric tighter and further reduce the deformation. Such a mandrel requires a collapsible design so that it can be inserted into the fabric tube easily.
In this work, explicit control of thread shape is not yet considered but this would further enhance the reliability of our system.

\section{Discussion and Conclusion}

In this paper, we have proposed a robotic system for manufacturing personalized stent grafts for AAA. Compared to other consumer goods, medical devices have a higher demand of service-orientation and customization. We have explored a practical solution for flexible production of personalized medical products at the system level. The modularized design increases the flexibility of the system and reduces the complexity of the task. For sewing different stent grafts, only the mandrel is required to be changed. This system can also be extended to other manufacturing tasks.

Experiments were conducted to evaluate the entire system in terms of its accuracy and robustness for sewing personalized stent grafts.
The experiments showed that this system presents sub-millimeter accuracy of positioning and for multiple throw sewing, it is able to achieve 79$\%$ of overall success. The targeted stitch size was 4.10 $mm$ and the system achieves an average stitch size of 3.60 $mm$.

In summary, the proposed system demonstrates a good potential for practical use and its performance can be improved by various approaches.
The focus of the future work will be to further increase the accuracy and robustness of the system. We will also further explore the flexibility of this system in sewing more complex stent grafts with fenestrations and branches.

\bibliographystyle{IEEEtran}
\bibliography{TII17}

\begin{IEEEbiography}
[{\includegraphics[width=1in,height=1.25in,keepaspectratio]{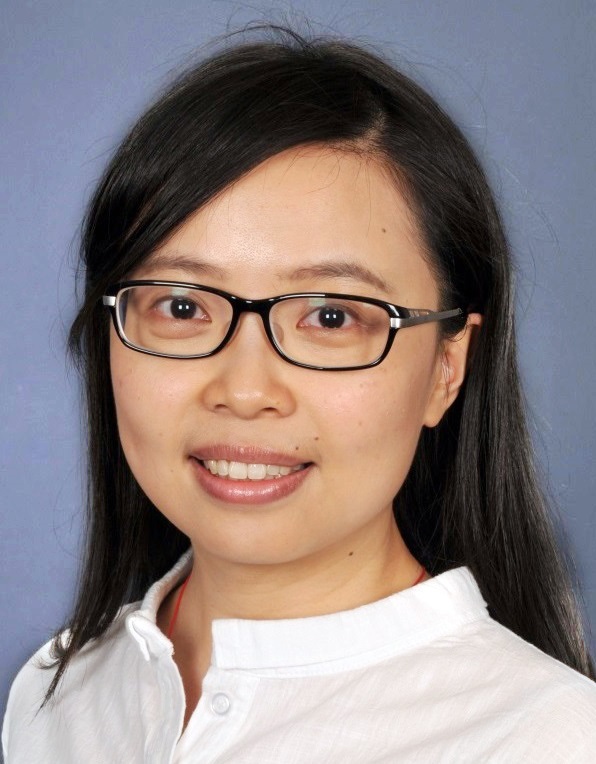}}]
{Bidan Huang} is currently a research associate at the Hamlyn Centre for Robotic Surgery, Imperial College London. She received her PhD degree in the University of Bath for her study on robotics in 2015. In 2012-2014, she was a visiting student of the Learning Algorithms and Systems Laboratory (LASA), Swiss Federal Institute of Technology in Lausanne (EPFL). Her main research interests are in robot learning, control, grasping and manipulation. Her research goal is to equip robot with human level hand dexterity for object manipulation in medical, industrial and daily life environments.  
\end{IEEEbiography}

\begin{IEEEbiography}
[{\includegraphics[height=32mm,keepaspectratio]{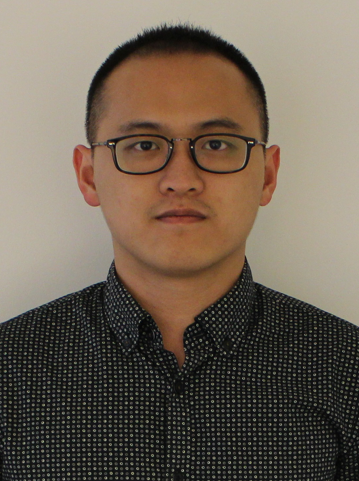}}]{Menglong Ye}  is currently a Computer Vision Scientist at Auris Surgical Robotics, Inc. He received a PhD degree in Surgical Vision and Navigation, and a Master’s degree in Medical Robotics, both from Imperial College London. He worked as a postdoc researcher at the Hamlyn Centre for Robotic Surgery from 2016 to 2017. His  research focus is placed on Surgical Vision and Navigation, aiming at developing computer vision techniques for assisting endoscopic navigation and surgical automation. He was previously funded by Institute of Global Health Innovation PhD Scholarship.
\end{IEEEbiography}


\begin{IEEEbiography}
[{\includegraphics[height=32mm,keepaspectratio]{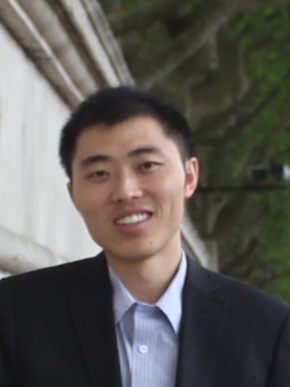}}]
{Yang Hu} received the B.S degree in Electronic Information Science and Technology from the College of Electronic Science and Engineering, Jilin University, China. He received the Master of Research degree in Surgical Robotics and Image Guided Intervention from the Hamlyn Centre, Imperial College London. Currently, he is a PhD candidate of the Hamlyn Centre, Imperial College London. His research interests include mechanical design, robot control, motion planning, and especially their medical/surgical applications. 
\end{IEEEbiography}

\begin{IEEEbiography}
[{\includegraphics[height=32mm,keepaspectratio]{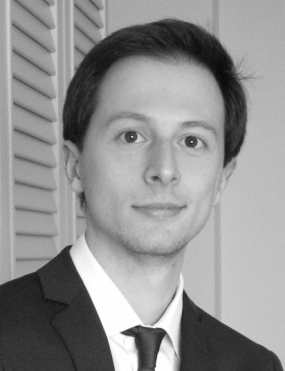}}]
{Alessandro Vandini} joined Samsung R\&D Institute UK, as a software engineer in 2017. 
He obtained his Ph.D. at the Hamlyn Centre for Robotic Surgery, Imperial College London. His Ph.D. research focused on surgical vision applied to endovascular and endoscopic procedures. This included tracking of surgical tools in intraoperative imaging modalities and vision-based shape sensing of surgical continuum robots. 
He received his Bachelor and Master degree in Computer Engineering from University of Modena and Reggio Emilia, Italy in 2008 and 2011, respectively.
\end{IEEEbiography}

\begin{IEEEbiography}
[{\includegraphics[height=32mm,keepaspectratio]{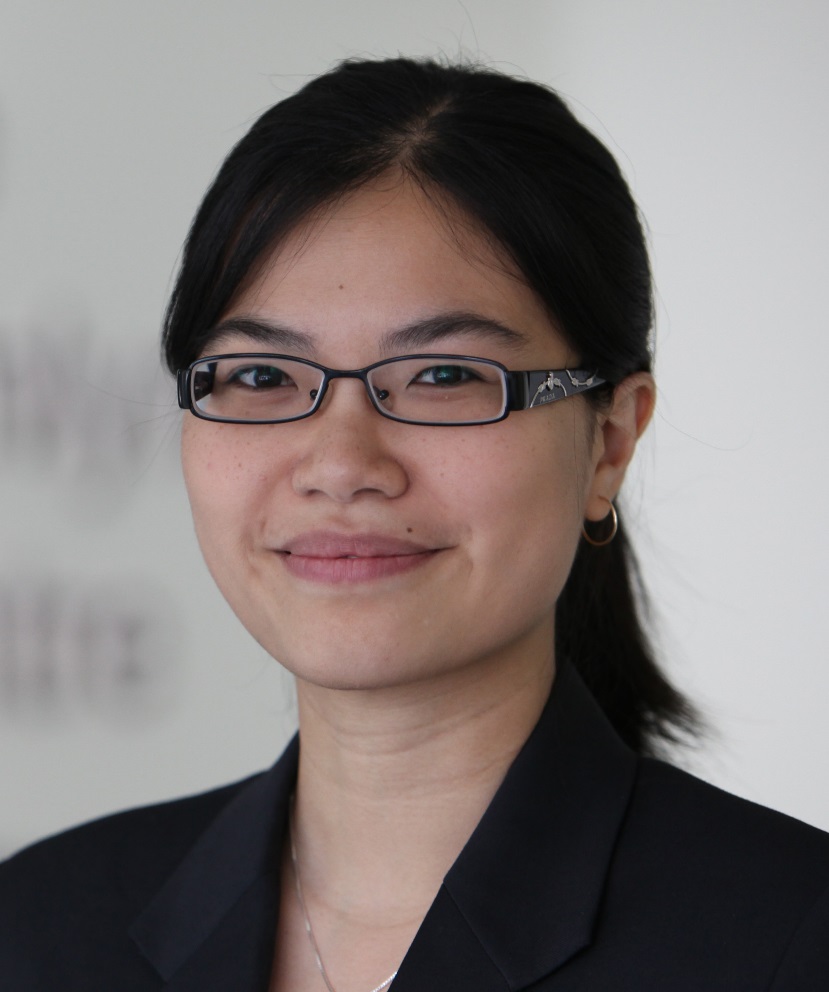}}]
{Su-Lin Lee} received her MEng. degree in information systems engineering and Ph.D. degree from Imperial College London, London, U.K., in 2002 and 2006, respectively, for her work on statistical shape modelling and biomechanical modelling.
She is currently a Lecturer at The Hamlyn Centre and Department of Computing, Imperial College London. Her main research interests are in computer assisted interventions, particularly for safer and more efficient robotic-assisted cardiovascular procedures.
\end{IEEEbiography}

\begin{IEEEbiography}
[{\includegraphics[height=32mm,keepaspectratio]{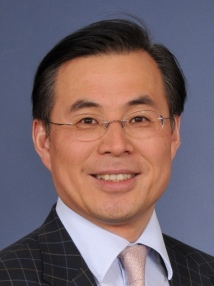}}]
{Guang-Zhong Yang} is director and co-founder of the Hamlyn Centre for Robotic Surgery. Professor Yang (FREng, FIEEE) is also the Chairman of the UK-RAS Network (http://ukras.org).  The mission of the UK-RAS Network is to provide academic leadership in RAS, expand collaboration with industry and integrate and coordinate activities of the EPSRC funded RAS capital facilities, Centres for Doctoral Training (CDTs) and partner universities across the UK.  Professor Yang's main research interests are in medical imaging, sensing and robotics. He is a Fellow of the Royal Academy of Engineering, fellow of IEEE, IET, AIMBE, IAMBE, MICCAI, CGI and a recipient of the Royal Society Research Merit Award and listed in The Times Eureka ``Top 100'' in British Science. He was awarded a CBE in the Queen's 2017 New Year Honour for his contribution to biomedical engineering. 
\end{IEEEbiography}

\end{document}